\documentclass[letterpaper, 10 pt, journal, twoside]{IEEEtran}

\usepackage{hyperref}
\usepackage{nameref}
\usepackage{subcaption}
\usepackage{placeins}
\usepackage{amsmath,amssymb,amsfonts}
\usepackage{tabularx}
\usepackage{pifont} % For checkmark symbols
\usepackage{footnote}

\usepackage{lipsum}
\usepackage[disable]{todonotes} %disable

\usepackage[english]{babel} % Load the babel package with your desired language
\addto\captionsenglish{} % Change the table name
\renewcommand{\thetable}{\arabic{table}}
\addto\captionsenglish{} % Change the table name

\newcommand{\eq}{Eq.~}
\newcommand{\fig}{Fig.~}
\newcommand{\tab}{Tab.~}
\newcommand{\sect}{Sec.~}

\newcommand{\rf}{F}
\newcommand{\kin}{\mathbf{k}^\mathrm{for}}
\newcommand{\coll}{W}

\newcommand{\rp}{RP-}

\newcommand{\rev}[1]{{#1}} %\newcommand{\rev}[1]{{\color{red} #1}}
\newcommand{\revv}[1]{{#1}}%\newcommand{\revv}[1]{{\color{red} #1}}

\DeclareMathOperator*{\argmax}{arg\,max}
\DeclareMathOperator*{\argmin}{arg\,min}

\renewcommand{\baselinestretch}{1}

\begin{document}
\bstctlcite{IEEEexample:BSTcontrol}

\title{Underactuated dexterous robotic grasping \\ with reconfigurable passive joints}
% \title{Dexterous Robotic Grasping with Reconfigurable Passive Joints}

\author{Marek~Kopicki$^{1}$, Sainul Islam Ansary$^{1}$, Simone Tolomei$^{2}$, Franco Angelini$^{2}$, Manolo Garabini$^{2}$ and Piotr Skrzypczyński$^{1}$
	\thanks{Manuscript received: July 3, 2024; Revised September 26, 2024; Accepted October 26, 2024.}%Use only for final RAL version
	\thanks{This paper was recommended for publication by Editor Clement Gosselin upon evaluation of the Associate Editor and Reviewers’ comments.
		
	This research is part of the project No. 2021/43/P/ST6/01921 co-funded by the National Science Centre and the European Union's Horizon 2020 research and innovation programme under the Maria Skłodowska-Curie grant agreement no. 945339} %Use only for final RAL version
	\thanks{$^{1}$ Institute of Robotics and Machine Intelligence, Poznan University of Technology, Poland. E-mail: \textit{\{marek.kopicki, sainul.uit\}@gmail.com}}
	\thanks{$^{2}$ Centro di Ricerca “Enrico Piaggio”, and Dipartimento di Ingegneria dell’Informazione, Università di Pisa, Pisa, Italy.}
	%\thanks{$^{3}$ This research is part of the project No. 2021/43/P/ST6/01921 co-funded by the National Science Centre and the European Union's Horizon 2020 research and innovation programme under the Maria Skłodowska-Curie grant agreement no. 945339.% For the purpose of Open Access, the author has applied a CC-BY public copyright licence to any Author Accepted Manuscript (AAM) version arising from this submission.
	%}
	\thanks{Digital Object Identifier (DOI): see top of this page.}
}

\IEEEoverridecommandlockouts

%\markboth{IEEE Robotics and Automation Letters. Preprint Version. Accepted Month, Year}
%{FirstAuthorSurname \MakeLowercase{\textit{et al.}}: ShortTitle} 
\markboth{IEEE Robotics and Automation Letters. Preprint Version. Accepted October, 2024}%
{Kopicki \MakeLowercase{\textit{et al.}}: Underactuated dexterous robotic grasping with reconfigurable passive joints}

\maketitle

\begin{abstract}

% New proposed Abstract by Simone:
\rev{We introduce a novel reconfigurable passive joint (\rp joint), which has been implemented and tested on an underactuated three-finger robotic gripper.} \rp joint has no actuation, but instead it is lightweight and compact. It can be easily reconfigured by applying external forces and locked to perform complex dexterous manipulation tasks, but only after tension is applied to the connected tendon. \rev{Additionally}, we present an approach that allows learning dexterous grasps from single examples with underactuated grippers and automatically configures the \rp joints for dexterous manipulation. \rev{This is enhanced by integrating kinaesthetic contact optimization, which improves grasp performance even further. The proposed RP-joint gripper and grasp planner have been tested on over 370 grasps executed on 42 IKEA objects and on the YCB object dataset, achieving grasping success rates of 80\% and 87\%, on IKEA and YCB, respectively.}

\end{abstract}


\IEEEpeerreviewmaketitle

\section{Introduction}

\IEEEPARstart{T}{he} advancement of robotic grasping technology is important for enhancing the efficiency and versatility of automation systems, particularly in dynamic environments such as warehouses and logistics centers. In these settings, robots must handle a diverse array of objects, varying in size, shape, and weight, necessitating the development of grippers that are both flexible and robust. Traditional robotic grippers often struggle with this diversity, making trade-offs between dexterity, payload capacity, and cost.

Underactuated grippers, which use fewer actuators than DoF, offer a promising solution due to their inherent compliance and simplicity. 
However, planning algorithms struggle with underactuated grippers, as their interaction with objects generates complex contact dynamics, making planning and control still a significant challenge~\cite{bonilla2014grasping}.
Reconfigurability can also be exploited to enhance the applicability and dexterity of grippers. It is especially useful to increase the robustness and the payload of robotic grippers, but it often comes at the cost of increased mechanical complexity~\cite{tavakoli2013flexirigid, Hsu2017}.

\begin{figure}[ht]
	\vspace*{-0.2cm}
\begin{center}
	\begin{tabular}{c}%trim=left botm right top
		\includegraphics[height=2.2cm]{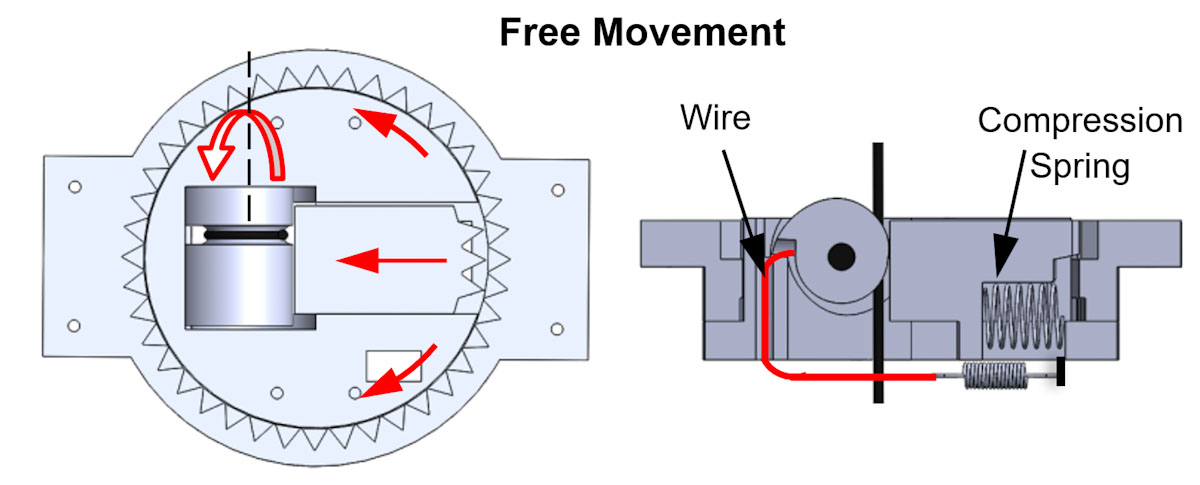}
		\includegraphics[height=2.2cm]{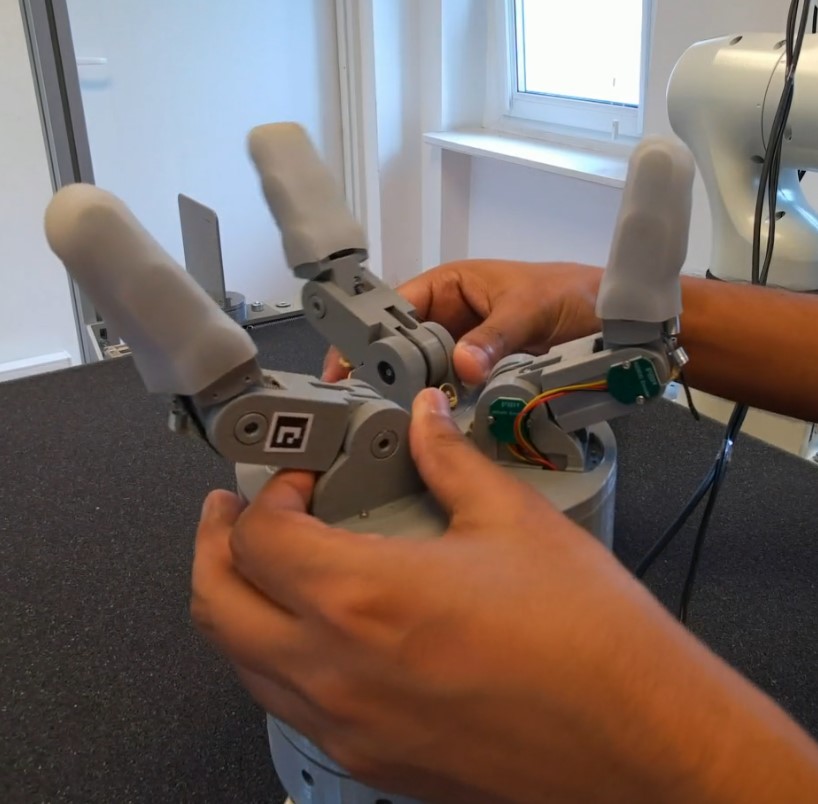}
	\end{tabular}\\
	
	\begin{tabular}{c}
		\includegraphics[height=2.2cm]{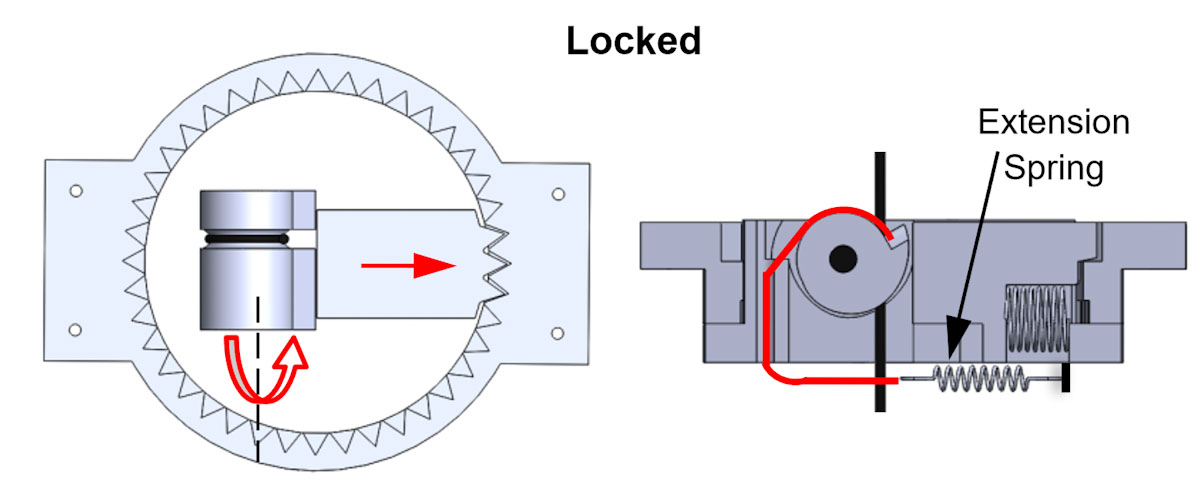}
		\includegraphics[height=2.2cm]{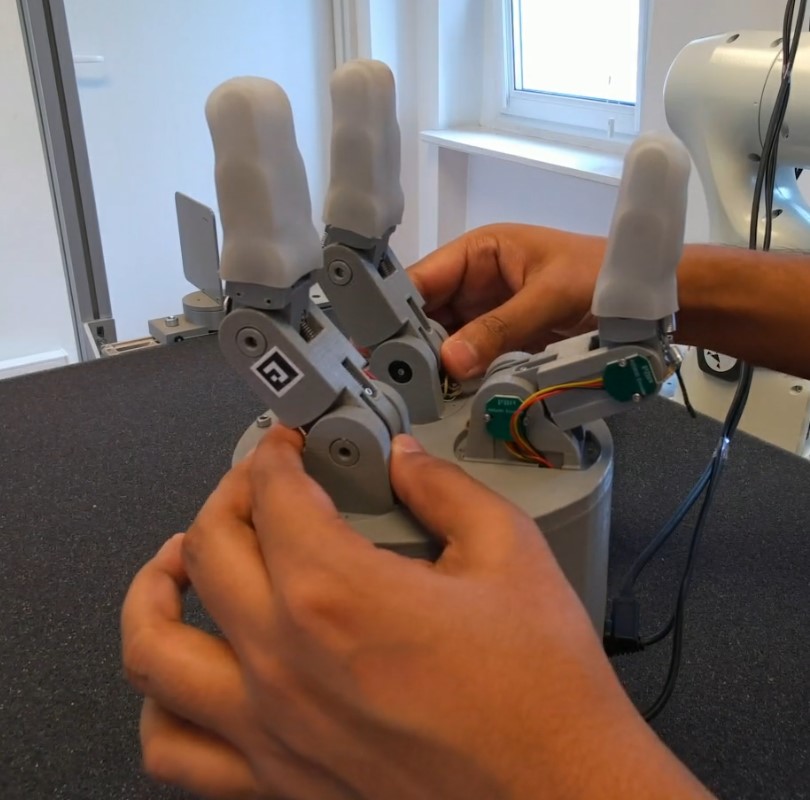}
	\end{tabular}    
\end{center}\vspace*{-0.3cm}

\caption{\rp joints are initially free to move (top row), then the tendon locks the joints, e.g. before grasping (bottom row). }
\label{fig:locking}\vspace*{-0.2cm}
\end{figure}

In this work, we designed an innovative underactuated and reconfigurable three-fingered gripper, that features high dexterity and applicability to different objects and scenarios. 
Our gripper features a novel reconfigurable passive joint (\rp joint, Fig. \ref{fig:locking}) mechanism that requires no internal actuation, simplifying the control system while enhancing the gripper's ability to conform to various object shapes and sizes. 
Furthermore, we developed a planner that elegantly resolves the complexities associated with grasp planning for this reconfigurable gripper. 
Our algorithm not only finds both grasp pose and gripper configuration, but also plans for interactions with the environment to change the gripper configuration.

%To address these challenges, it is essential to design grippers that can adapt to different objects while maintaining reliability and affordability. Underactuated grippers, which use fewer actuators than degrees of freedom, offer a promising solution due to their inherent compliance and simplicity. However, achieving the right balance between adaptability and control remains a significant problem.

%In this paper, we introduce an innovative underactuated three-fingered robotic gripper featuring a novel passive joint mechanism. Unlike conventional designs, our gripper's passive joints require no internal actuation, simplifying the control system while enhancing the gripper's ability to conform to various object shapes and sizes. 
% This design facilitates systematic in-hand manipulations, ensuring robust and flexible grasping capabilities suitable for a wide range of applications.

%The contributions of this work are threefold. First, we present the design and implementation of the \rp joint mechanism within an underactuated gripper. \rev{Second, we demonstrate the gripper's capabilities through experimental tests on the prototype.}
%Second, we analyze the theoretical grasping workspace and demonstrate the gripper's capabilities through both simulations and tests on the prototype.
%Finally, we provide algorithms for computing manipulation maps and planning manipulation tasks, showcasing the gripper's effectiveness with objects of different sizes and shapes chosen from the world-wide available IKEA items \rev{and from the YCB~\cite{calli2015ycb} dataset of objects}.
\rev{
The contributions of this work are twofold. First, we present the design and implementation of the \rp joint mechanism within an underactuated gripper. 
We then present a grasping algorithm able to both select grasp poses and to plan arm movement fully exploiting the \rp joints' characteristics. We perform experimental validation on the world-wide available IKEA items and from the YCB~\cite{calli2015ycb} dataset of objects, showcasing the gripper's effectiveness with objects of different sizes and shapes.}
%We then provide algorithms for computing manipulation maps and planning manipulation \rev{tasks that exploit \rp joints}, showcasing the gripper's effectiveness with objects of different sizes and shapes chosen from the world-wide available IKEA items \rev{and from the YCB~\cite{calli2015ycb} dataset of objects}. We then provide algorithm

%The remainder of this paper is structured as follows: Section II reviews related work in the field of robotic grasping and underactuated mechanisms. Section III details the design and kinematics of the proposed gripper. Section IV presents the theoretical analysis of the grasping workspace and manipulation capabilities. Section V discusses the implementation and experimental validation of the gripper. Finally, Section VI concludes the paper. 

\vspace*{-0.3cm}
\section{Related work}

\subsection{Gripper design}

%TODO: SOTA, comparison

Underactuation in gripper design is a common approach to lower system complexity~\cite{shintake2018soft} and increase robustness in grasping tasks~\cite{bonilla2014grasping}.
Common approaches to achieve underactuation in grippers include the use of continuously deformable materials in the gripper~\cite{deimel2013compliant, deimel2016novel, glick2018soft}, or the use of articulated compliant mechanisms~\cite{odhner2014compliant}.
The latter approach can result in multi-modal grippers~\cite{liu2023hybrid} and anthropomorphic hands like the Pisa/IIT SoftHand~\cite{della2018toward}.
With a similar approach,~\cite{angelini2020softhandler} presents a soft underactuated four-fingered gripper that allows for robust grasping, especially with fragile or irregular objects.

Reconfigurability has emerged as a key feature in modern underactuated gripper designs. In~\cite{Lu2021}, it is presented a reconfigurable underactuated robot hand excelling in systematic prehensile in-hand manipulations across various object sizes and shapes, featuring a two degrees of freedom (DoF) five-bar linkage palm with three underactuated fingers controlled by a single actuator. Similarly,~\cite{Chappel2023} introduced the Hydra Hand, a novel reconfigurable four-fingered underactuated robotic gripper achieving both compliant power and rigid precision grasps using a single motor and hydraulic actuator.
The advantages of these works are hindered by the increased mechanical complexity required to implement the ability to reconfigure.

In order to reduce design complexity, reconfigurability can also be achieved using locking mechanisms~\cite{Plooij2015}. Literature already proposes the integration of locking mechanisms with underactuated hands. \rev{However, most solutions aim at providing multi-mode/adaptive grasping~\cite{Hermann2019}, firm/secure grasps~\cite{Hsu2017}, increasing object graspability~\cite{tavakoli2013flexirigid} or at reducing energy consumption in presence of high payloads.}
% However, most solutions aim at providing firm and adaptive grasps or at reducing energy consumption in presence of high payloads. In~\cite{Hermann2019}, the Authors developed a dual-mode robotic gripper incorporating both an underactuated grasping mode and a fully actuated precision mode, using electromagnetically frictional locking joints to switch between modes. This approach allows for shape conformity in grasping and enhanced payload handling through individual joint locking.
% The SLUM gripper in~\cite{Hsu2017} combines underactuation with a frictional locking mechanism, demonstrating the ability to lock a grasp securely and support significant payloads. 
% In~\cite{tavakoli2013flexirigid}, it is proposed a 2-fingers gripper that can lock together the finger tips to increase object graspability.
To the best of \revv{the a}uthors' knowledge, the application of locking mechanisms to underactuated hands with the goal of enhancing reconfigurability is understudied, and it is usually limited to lockable finger joints to modify hand closure, such as in~\cite{mitsui2013under} using latch mechanism and in~\cite{peerdeman2013development} using self-amplifying brakes.

In this work, we propose the use of a latch locking mechanism at the fingers base to enable the modification of the gripper overall configuration. This leads to the execution of a variety of grasps with a small number of actuators, and therefore to an increased gripper dexterity and the applicability to different grasping scenarios. \rev{For example, a single \rp joint could be mounted at the base of the thumb in a prosthetic hand with one synergy providing greater grip strength than in hands with more synergies. A patient with an artificial thumb or wrist could swiftly reconfigure the thumb prior to manipulation, while still firmly locking it when large and unpredictable forces are applied.}

\vspace*{-0.4cm}
\subsection{Grasping with underactuated grippers and hands}
% Intro of general grasping problem
Grasping unknown objects, despite being a popular research topic~\cite{newbury2023deep}, still remains a challenging task, as even current approaches are outperformed by human abilities~\cite{billard2019trends}.
Moreover, grasping with underactuated grippers makes the problem even more challenging as it inherits the issues associated with underactuated control.

On fully actuated grippers, analytical solutions to the grasping problem often rely on contact dynamics theory~\cite{bicchi2000robotic, pollayil2022sequential}, which implies having a detailed geometrical model of the object, often hard to obtain in real-world tasks.
This has lately driven the research effort towards learning-based methodologies~\cite{kleeberger2020survey}, which exploit either simulated data ~\cite{mahler2019learning} or real-world grasping data~\cite{levine2018learning}. 
In particular,~\cite{mahler2019learning} proposes an architecture based on a Convolutional Neural Network (CNN) trained on synthetic data to select grasping points. 
%Similarly, ~\cite{ten2017grasp} uses a CNN to predict the quality of a grasp from depth images, and~\cite{ainetter2021end} uses semantic segmentation from a CNN to predict grasping candidates.
Similarly, ~\cite{ten2017grasp} uses a CNN to predict the quality of a grasp from depth images, and~\cite{ainetter2021end} uses semantic segmentation to find grasping candidates.
Finally,~\cite{Aktas2022Deep} exploits CNNs to learn generative and evaluative models, able to respectively predict and rank grasp poses.
\rev{More recently, \cite{urain2023se} uses diffusion models to learn joint motion and grasp poses from a simulated dataset of 90k samples.}

Learning from examples also has been proven effective. In~\cite{kopicki2014learning}, we showed that using contact models it is possible to learn highly-generic dexterous grasp strategies that can easily transfer across different categories of objects, without any object models. The method was further extended to enable grasping of objects seen from a single viewpoint~\cite{kopicki2016oneshot}\cite{kopicki2019learning}, and compressing improved contact models to enable self-learning from experience ~\cite{kopicki2019learning}.
In this letter, we present a planning algorithm that extends upon our previous work making it robust to the challenges of underactuated grippers and enabling planning for reconfigurability of the gripper itself.

%TODO: Extend a bit,
Data-driven techniques have also been proven effective when dealing with underactuated grippers.
Notably,~\cite{della2019learning} trained a classifier from human demonstration to select one among a set of human-inspired grasping primitives.
Similarly,~\cite{palleschi2023grasp} uses a bounding box decomposition of objects and a decision tree learned from human demonstrations to determine a suitable grasp pose either with a fully actuated or an underactuated gripper.
Despite being effective,~\cite{palleschi2023grasp} relies on a dataset of demonstrations of humans grasping objects using the specific robotic hand.
In contrast to these works, our planner is able to find grasp poses for our underactuated gripper using a very limited set of demonstrations.

\rev{A multitude of methods and algorithms for grasping, along with varying gripper designs and diverse datasets or benchmarks, makes direct comparison both complex and challenging. To maintain a concise and fair overview, we summarize the related work in \tab\ref{tab:comparison} highlighting key features of the referenced papers and situating our work in this context.}

\begin{table}[h]
\renewcommand{\arraystretch}{1.1} % Adjusts row height slightly for readability
\setlength{\tabcolsep}{2pt} % Adjusts the space between columns
\rev{
% \begin{tabularx}{\columnwidth}{|X||X|X|X|X|X|X|X|}
\begin{tabularx}{\columnwidth}{|>{\centering\arraybackslash}X||>{\centering\arraybackslash}X|>{\centering\arraybackslash}X|>{\centering\arraybackslash}X|>{\centering\arraybackslash}X|>{\centering\arraybackslash}X|>{\centering\arraybackslash}X|>{\centering\arraybackslash}X|}
\hline
\textbf{Paper \newline Ref.} & \textbf{Dext. \newline Grasp.} & \textbf{Under- \newline actuat.} & \textbf{Model \newline Free}& \textbf{Novel \newline Objects} & \textbf{Single \newline View} & \textbf{Grasp Success}\\
\hline
\cite{mahler2019learning}         &          &          &          &          &\ding{51} &95\%   \\
\cite{ten2017grasp}               &          &          &          &          &\ding{51} &93\%   \\
\cite{ainetter2021end}            &          &          &\ding{51} &\ding{51} &\ding{51} &89\%   \\
\cite{Aktas2022Deep}              &\ding{51} &          &\ding{51} &\ding{51} &\ding{51} &88\%   \\
%\cite{urain2023se} (?)           &          &          &          &          &          &     \\
\cite{kopicki2014learning}        &\ding{51} &          &          &\ding{51} &\ding{51} &86\%  \\
%\cite{kopicki2016oneshot}        &\ding{51} &          &\ding{51} &\ding{51} &\ding{51} &     \\
\cite{kopicki2019learning}        &\ding{51} &          &\ding{51} &\ding{51} &\ding{51} &88\%   \\
\cite{della2019learning}          &\ding{51} &\ding{51} &          &          &          &81\%   \\
\cite{palleschi2023grasp}         &\ding{51} &\ding{51} &          &\ding{51} &          &94\%   \\   
\cite{bekiroglu2019benchmarking}  &          &          &\ding{51} &\ding{51} &          &95\%   \\
IKEA                             &\ding{51} &\ding{51} &\ding{51} &\ding{51} &\ding{51} &80\%   \\
YCB                              &\ding{51} &\ding{51} &\ding{51} &\ding{51} &\ding{51} &87\%   \\
\hline
\end{tabularx}
\caption{\rev{Summary of key features in related work on grasping techniques and grippers. IKEA and YCB are for our work.}}
% CHECK MARKS ARE TENTATIVE
\label{tab:comparison}
}
\end{table}
\vspace*{-0.8cm}

%\cite{mahler2019learning}  uses instance recognition, so it no to novel objects
%\cite{ten2017grasp} uses known CAD models to detect grasp pose
%\cite{ainetter2021end} RGB, "Note that for grasp detection of unseen objects, we define the set of semantic classes as {graspable, non-graspable}", so likely novel objects
%\cite{della2019learning} RGB, 
% \cite{bekiroglu2019benchmarking} used a model-free and learning-free grasp planner making all objects novel to it and one-shot not applicable. 
% Here we start the paragraph with quantitative compariosn of the papers based on a table. 

\section{Underactuated gripper with \rp joints}
\begin{figure}[ht]
\vspace*{-0.3cm}
\begin{tabular}{ccc}%trim=left botm right top
\begin{tabular}{c}
\hspace*{-0.4cm}\includegraphics[height=3.8cm, trim=0.0cm 0.0cm 0.0cm 0.0cm,clip]{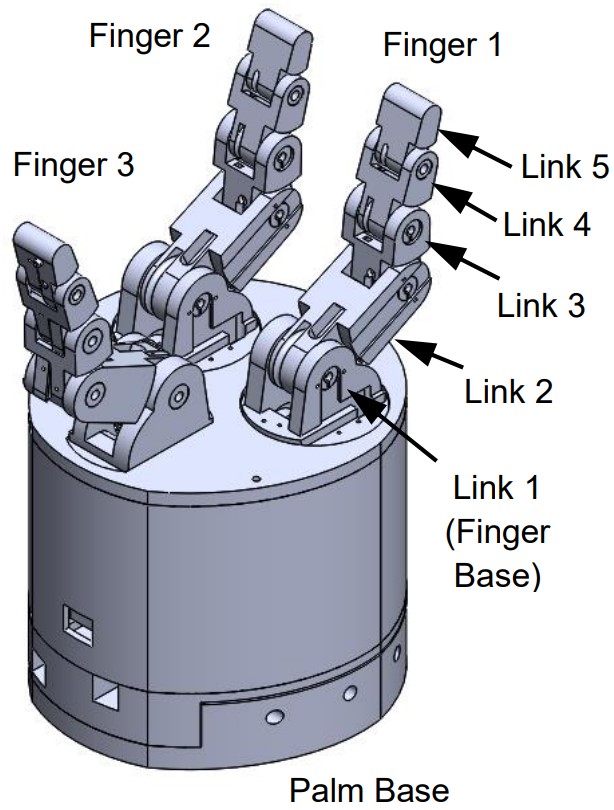}\\\hspace*{-1.0cm}
\begin{minipage}{0cm}\subcaption{}\label{fig:hand_a}\end{minipage}
\end{tabular}&

\begin{tabular}{c}
\hspace*{-0.8cm}\includegraphics[height=3.8cm, trim=0.0cm 0.0cm 0.0cm 0.0cm,clip]{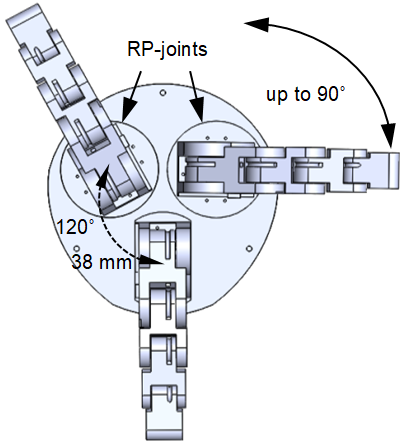}\\\hspace*{-1.0cm}
\begin{minipage}{0cm}\subcaption{}\label{fig:hand_b}\end{minipage}
\end{tabular}&

\begin{tabular}{c}
\hspace*{-0.7cm}\includegraphics[height=1.7cm, trim=0.0cm 0.0cm 0.0cm 1.0cm,clip]{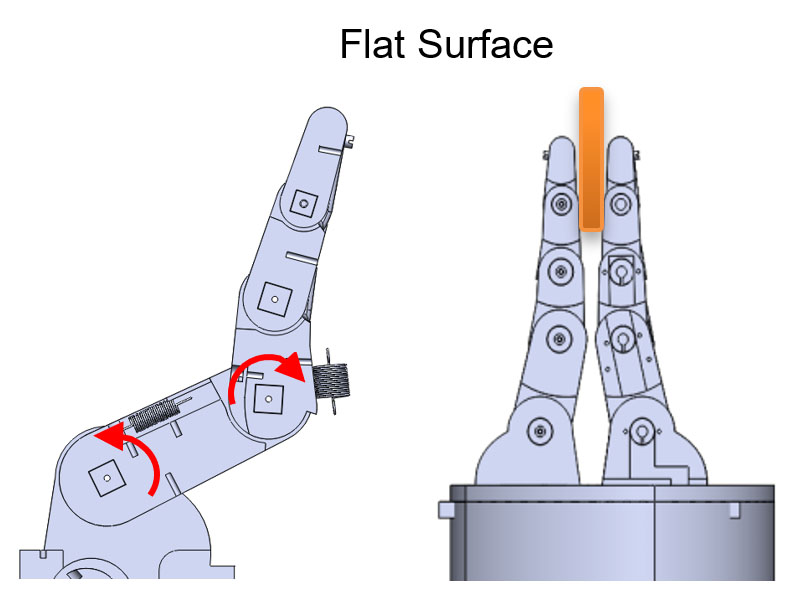}\vspace*{-0.3cm}
\\\hspace*{-1.0cm}\begin{minipage}{0cm}\subcaption{}\label{fig:tendon_c}\end{minipage}\\
\hspace*{-0.9cm}\includegraphics[height=1.7cm, trim=0.0cm 0.0cm 0.0cm 1.0cm,clip]{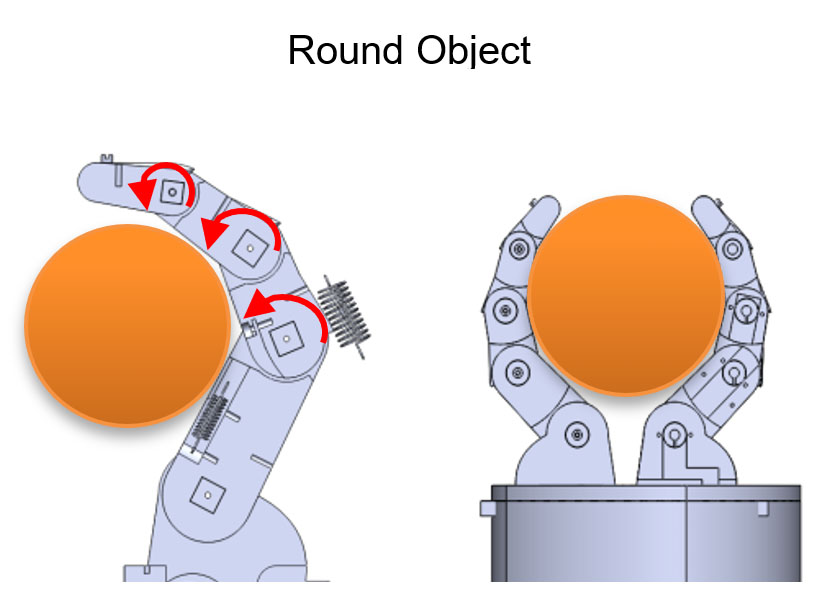}\vspace*{-0.0cm}
\\\hspace*{-1.0cm}\begin{minipage}{0cm}\subcaption{}\label{fig:tendon_d}\end{minipage}
\end{tabular}
\end{tabular}

\caption{\rev{The underactuated gripper (a), its finger configurations (b), finger adaptation to flat (c) and round (d) object.}}
\label{fig:hand}
\end{figure}

\rev{The gripper (\fig\ref{fig:hand_a}) consists of three kinematically identical fingers placed on a palm-base with a circular pattern (radius 38 mm and 120\revv{\textdegree}~apart \fig\ref{fig:hand_b}).} The design is inspired by the dexterity of the human hand, features two adjustable fingers and one stronger fixed finger, mimicking the functionality of an opposable thumb. Each finger has five links connected by active and (reconfigurable) passive joints. \rev{The} \textit{\rp joint} has no direct action of its own, but it can be controlled by external forces. \rev{The link lengths are 26, 45, 25, 25, and 20 mm, respectively. The gripper has an overall mass of 978 grams.}
%The passive joints make the gripper reconfigurable in the sense that its fingers can be rotated by applying external forces about the palm-base to allow grasping of objects of varied shapes like cylindrical, spherical, etc.
\rp joints allow the gripper to be reconfigured so that its fingers could be rotated around the base by applying external forces to improve the gripper's dexterity to reliably grip objects with a wider range of shapes (\fig\ref{fig:hand_b}).
The \textit{active joints} are directly controlled/actuated by tendon wires.

% \begin{figure}[ht]

% \begin{tabular}{ccc}%trim=left botm right top
% \begin{tabular}{c}
% \hspace*{-0.3cm}\includegraphics[height=3.8cm, trim=0.0cm 0.0cm 0.0cm 0.0cm,clip]{resources/Gripper_a.jpg}\\\hspace*{-1.0cm}
% \begin{minipage}{0cm}\subcaption{}\label{fig:hand_a}\end{minipage}
% \end{tabular}&

% \begin{tabular}{c}
% \hspace*{-0.8cm}\includegraphics[height=3.8cm, trim=0.0cm 0.0cm 0.0cm 0.0cm,clip]{resources/Gripper_b.jpg}\\\hspace*{-1.0cm}
% \begin{minipage}{0cm}\subcaption{}\label{fig:hand_b}\end{minipage}
% \end{tabular}&

% \begin{tabular}{c}
% \hspace*{-0.8cm}\includegraphics[height=3.8cm, trim=0.0cm 0.0cm 0.0cm 0.0cm,clip]{resources/Gripper_d.jpg}\\\hspace*{-1.0cm}
% \begin{minipage}{0cm}\subcaption{}\label{fig:hand_c}\end{minipage}
% \end{tabular}
% \end{tabular}

% \caption{The three-finger underactuated griper and its different finger configurations.}
% \label{fig:hand}
% \end{figure}\vspace*{-0.4cm}

\vspace*{-0.7cm}
\subsection{Tendon driven actuation mechanism}

% \begin{figure}[ht]

% \begin{center}
% \begin{tabular}{ccc}%trim=left botm right top
% \begin{tabular}{c}
% \hspace*{-0.5cm}\includegraphics[height=3.8cm, trim=0.0cm 0.0cm 0.0cm 0.0cm,clip]{resources/Tendon_a.jpg}\\\hspace*{-0.0cm}
% \begin{minipage}{0cm}\subcaption{}\label{fig:tendon_a}\end{minipage}
% \end{tabular}&

% \begin{tabular}{c}
% \hspace*{-0.5cm}\includegraphics[height=3.8cm, trim=0.0cm 0.0cm 0.0cm 0.0cm,clip]{resources/Tendon_b.jpg}\\\hspace*{-0.0cm}
% \begin{minipage}{0cm}\subcaption{}\label{fig:tendon_b}\end{minipage}
% \end{tabular}&

% \begin{tabular}{c}
% \hspace*{-0.1cm}\includegraphics[height=1.7cm, trim=0.0cm 0.0cm 0.0cm 1.0cm,clip]{resources/Adaptation_a.jpg}\vspace*{-0.3cm}
% \\\hspace*{-1.0cm}\begin{minipage}{0cm}\subcaption{}\label{fig:tendon_c}\end{minipage}\\
% \hspace*{-0.1cm}\includegraphics[height=1.7cm, trim=0.0cm 0.0cm 0.0cm 1.0cm,clip]{resources/Adaptation_b.jpg}\vspace*{-0.0cm}
% \\\hspace*{-1.0cm}\begin{minipage}{0cm}\subcaption{}\label{fig:tendon_d}\end{minipage}
% \end{tabular}
% \end{tabular}
% \end{center}\vspace*{-0.3cm}

% \caption{Tendon routing inside finger (a), four-bar linkage mechanism (b), finger adaptation to flat (c) and round (d) objects.}
% \label{fig:tendon}\vspace*{-0.3cm}
% \end{figure}

\begin{figure}[ht]
\vspace*{-0.3cm}
\begin{center}
\begin{tabular}{cc}%trim=left botm right top
\begin{tabular}{c}
\hspace*{-0.5cm}\includegraphics[height=4.8cm, trim=0.0cm 0.0cm 0.0cm 0.0cm,clip]{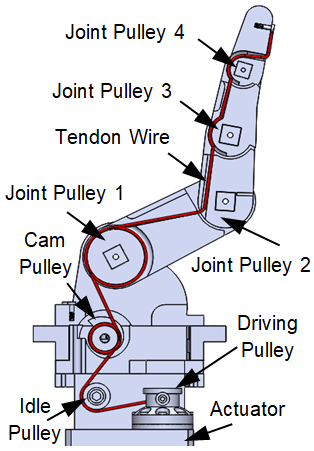}\hspace*{0.8cm}
% \begin{minipage}{0cm}\subcaption{}\label{fig:tendon_a}\end{minipage}
\end{tabular}&

\begin{tabular}{c}
\hspace*{-0.5cm}\includegraphics[height=4.8cm, trim=0.0cm 0.0cm 0.0cm 0.0cm,clip]{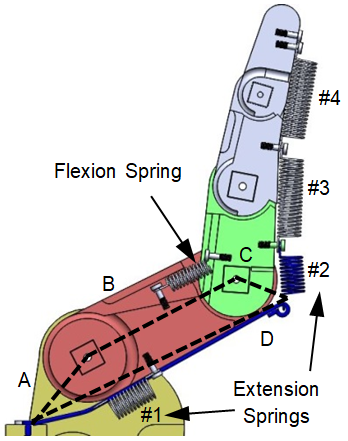}
% \begin{minipage}{0cm}\subcaption{}\label{fig:tendon_b}\end{minipage}
\end{tabular}
\end{tabular}
\end{center}\vspace*{-0.3cm}

\caption{\rev{Tendon routing inside finger (left) and four-bar linkage mechanism (right).}}
\label{fig:tendon}\vspace*{-0.3cm}
\end{figure}

A single tendon wire in each finger, routed over joint pulleys, transfers power from a servo actuator located in the palm-base (\fig\ref{fig:tendon}). The tendon wires provide flexion motion, while linear extension springs at each joint facilitate extension. The tendon routing, illustrated in \fig\ref{fig:tendon} (left), starts from a driving pulley attached to the servo motor and proceeds over idle and cam pulleys, and joint pulleys, finally fixing to the distal link. This configuration enables the gripper to adapt its fingers to objects' surfaces using a four-bar linkage mechanism and optimized joint pulley diameters, tendon routing, and spring stiffness (\fig\ref{fig:tendon} right). \rev{The four-bar linkage is formed by the finger base (A, yellow), second finger-link (B, red), second joint-pulley (C, green) attached to the third link, and the second extension spring with a connecting wire (D, blue). The dashed lines represent the four-bar linkage mechanism.} This mechanism allows the finger to adapt to various object shapes, as shown in \fig\ref{fig:tendon_c} and \fig\ref{fig:tendon_d}. The mechanism helps the gripper to switch between parallel and enveloping grasps inspired by \rev{the} velo gripper~\cite{ciocarlie2014velo}, but the mechanism is not the same. Unlike the velo gripper which used a separate constraint tendon, here the spring-loaded wire (D) acts both as the constraint needed to keep fingers parallel as well as the extensor for the second joint. This and the design of the second joint reduce the overall complexity.

\begin{figure}[ht]
\begin{center}
\vspace*{-0.2cm}
\begin{tabular}{cccc}%trim=left botm right top
\begin{tabular}{c}
\hspace*{-0.4cm}\includegraphics[height=2.5cm, trim=0.0cm 0.0cm 0.0cm 0.0cm,clip]{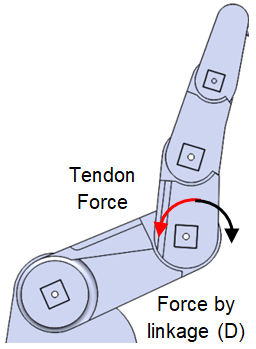}\\\hspace*{-1.3cm}
\begin{minipage}{0cm}\subcaption{}\label{fig:compliant_a}\end{minipage}
\end{tabular}&

\begin{tabular}{c}
\hspace*{-0.5cm}\includegraphics[height=2.5cm, trim=0.0cm 0.0cm 0.0cm 0.0cm,clip]{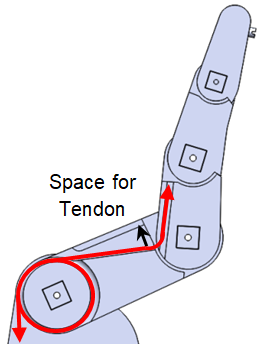}\\\hspace*{-2.0cm}
\begin{minipage}{0cm}\subcaption{}\label{fig:compliant_b}\end{minipage}
\end{tabular}&

\begin{tabular}{c}
\hspace*{-0.5cm}\includegraphics[height=2.5cm, trim=0.0cm 0.0cm 0.0cm 0.0cm,clip]{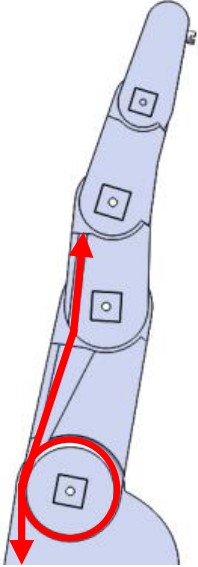}\\\hspace*{-1.2cm}
\begin{minipage}{0cm}\subcaption{}\label{fig:compliant_c}\end{minipage}
\end{tabular}&

\begin{tabular}{c}
\hspace*{-0.4cm}\includegraphics[height=2.5cm, trim=0.0cm 0.0cm 0.0cm 0.0cm,clip]{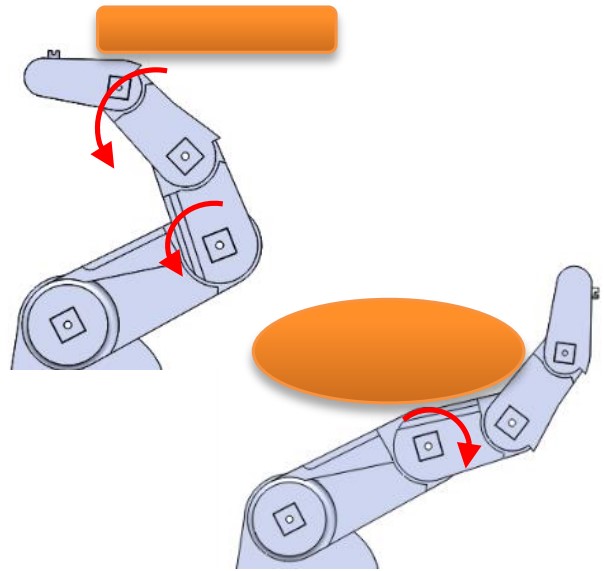}\\\hspace*{-1.0cm}
\begin{minipage}{0cm}\subcaption{}\label{fig:compliant_d}\end{minipage}
\end{tabular}
\end{tabular}
\end{center}
\vspace*{-0.3cm}
\caption{\rev{Design of the second joint.}}
\label{fig:compliant}\vspace*{-0.3cm}
\end{figure}

%\textit{Second Joint Design}. 
The second joint is crucial for achieving parallel and enveloping grasps. It must move in both directions to function correctly. To prevent the joint from locking due to opposing forces, the second pulley has a non-linear moment arm that adjusts the tendon torque based on joint position, as depicted in \fig\ref{fig:compliant}. Initially, the tendon produces minimal forward torque, allowing the four-bar linkage to move the joint backward. Once the finger reaches a perpendicular position (\fig\ref{fig:compliant_c}), increased tendon torque moves the second joint forward, facilitating an enveloping grasp. This design also provides physical compliance, enhancing gripper safety by allowing the last three links to move in both directions under external forces, preventing collision damage (\fig\ref{fig:compliant_d}).

\subsection{\rp joint mechanism}

%The idea of passive joins is to allow the fingers to spread sideways freely during pre-grasp gripper configuration and lock itself during grasping. The proposed mechanism provides two extra degrees of freedom to the gripper without employing any direct actuators. The only drawback is that the joint movement depends on the external forces. The working principle of the passive joint is shown in \fig\ref{fig:passive}. 

\begin{figure}[ht]
\vspace*{-0.2cm}
\begin{center}
\includegraphics[height=3cm, trim=0.0cm 0.0cm 0.0cm 0.0cm,clip]{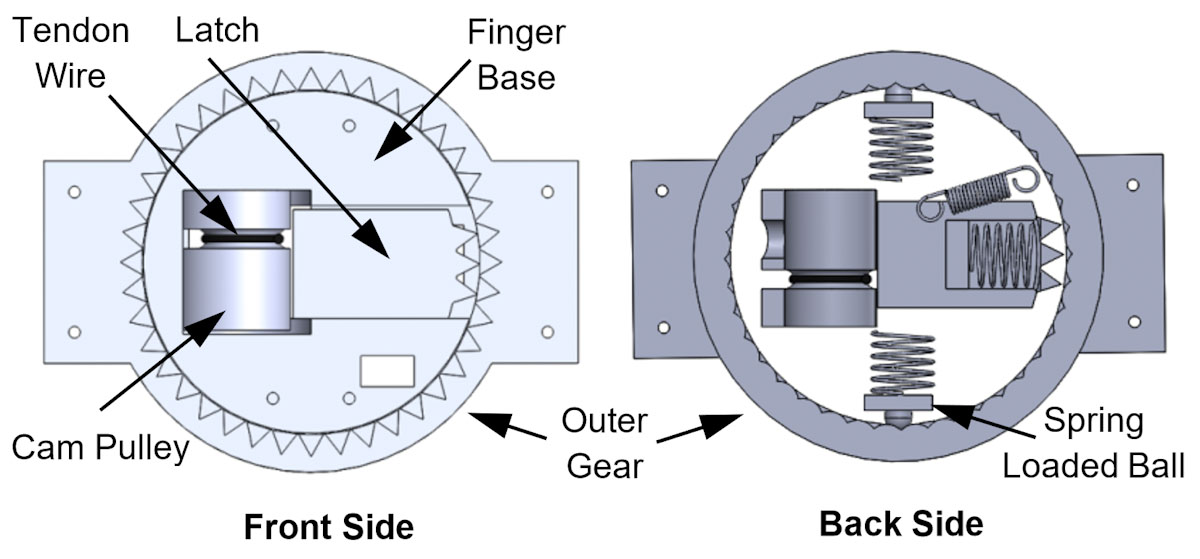}
\end{center}
\vspace*{-0.3cm}
\caption{\rp joint mechanism}\label{fig:passive}
\vspace*{-0.3cm}
\end{figure}

The idea behind \rp joints is to increase the dexterity of the gripper in manipulation scenarios involving the handling of objects with complex shapes. The proposed mechanism provides two additional DoF for the gripper without using any direct actuators. The main disadvantage is that changing the position of the joint requires the use of external forces. The principle of operation of the \rp joint is shown in \fig\ref{fig:passive}.

The \rp joint consists of an outer gear, a finger base, a cam pulley and a spring loaded latch. The outer gear is fixed on the palm base and together with the finger base form a revolute joint. The first link of a finger is fixed to the finger base and can rotate freely under external force. 
The outer gear has two layers of teeth on the front and back sides of the gear with different tooth profiles serving two purposes as shown in \fig\ref{fig:passive}. Firstly, the frontal side with triangular teeth and the latch with same tooth-profile do the joint locking during grasping.
The sliding latch locks the joint once pushed by the cam pulley. The cam pulley is driven by a tendon which is used for actuating the rest of the finger joints. The tendon has no direct role in controlling the \rp joint motion and it only helps to lock/unlock the joint. The \rp joint works in two stages i.e., free movement and locking. In free movement stage as shown in \fig\ref{fig:locking} (top), the cam pulley and latch remain at retract position, which allows the finger base or the attached finger to rotate freely whenever an external force applied on the finger. In locked stage, this happens only during the object grasping when the actuator applies tension to the tendon (\fig\ref{fig:locking} (bottom)). The tendon rotates the cam pulley which pushes the latch to lock the finger base with the outer gear. The free movement stage is realised when the actuator releases the tendon. Now at this stage, the tendon fails to fully retract the cam pulley as the tendon loses its tension. So, this problem is overcome by an extension spring attached to the cam pulley using a small tendon wire as shown in \fig\ref{fig:locking}. A compression spring attached to the latch helps to retract the latch to its initial position while the cam pulley returns to its retract position leading to unlocking of the joint.

The back side with circular profile provides necessary friction as well as helps the latch and frontal gear to remain aligned. A small amount of friction is necessary to make the \rp joint stationary when no external force is applied on the force otherwise a complete free moving joint will be uncontrollable and also gravity can cause undesirable joint movements while robot arm moving. Two spring loaded balls are used for this purpose and the springs push and hold them against the circular profile of the outer gear to produce the desired friction and the alignment.

\vspace*{-0.3cm}

\subsection{Gripper design, sensorization, electronics, and control}\label{sec::design}
\begin{figure}[ht]
\vspace*{-0.2cm}
    \begin{center}
    \begin{tabular}{ccc}%trim=left botm right top
        % \begin{tabular}{c}
        % \hspace*{-0.4cm}\vspace*{3.0cm}\begin{minipage}{0cm}\subcaption{}\label{fig:assembly_a}\end{minipage}
        % \end{tabular}&
        \begin{tabular}{c}
        \hspace*{-0.4cm}\includegraphics[height=3.4cm, trim=0.0cm 0.0cm 0.0cm 0.0cm,clip]{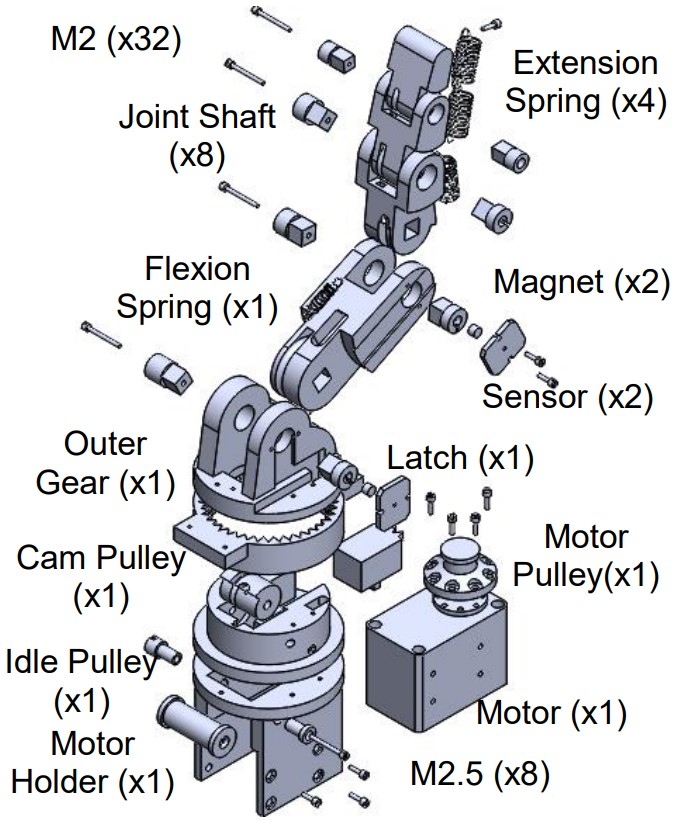}\\\hspace*{-1.0cm}
        % \begin{minipage}{0cm}\subcaption{}\label{fig:assembly_a}\end{minipage}
        \end{tabular}&

        % \begin{tabular}{c}
        % \hspace*{-0.4cm}\vspace*{3.0cm}\begin{minipage}{0cm}\subcaption{}\label{fig:assembly_b}\end{minipage}
        % \end{tabular}&
        
        \begin{tabular}{c}
        \hspace*{-0.7cm}\includegraphics[height=3.4cm, trim=0.0cm 0.0cm 0.0cm 0.0cm,clip]{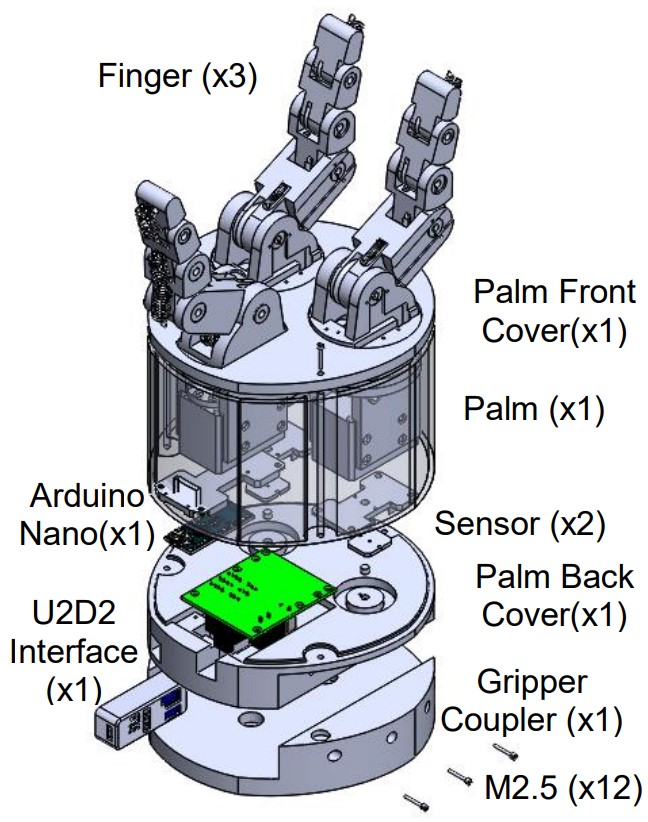}\\\hspace*{-1.0cm}
        % \begin{minipage}{0cm}\subcaption{}\label{fig:assembly_b}\end{minipage}
        \end{tabular}&
    
        \begin{tabular}{c}
        \hspace*{-0.6cm}\includegraphics[height=3.0cm, trim=0.0cm 0.0cm 0.0cm 0.0cm,clip]{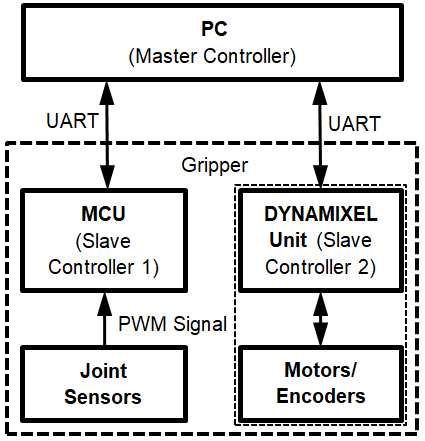}\\\hspace*{-1.0cm}
        % \begin{minipage}{0cm}\subcaption{}\label{fig:assembly_b}\end{minipage}
        \end{tabular}
    \end{tabular}
    \end{center}\vspace*{-0.6cm}
    
    \caption{\rev{Exploded views of gripper assemblies and controller.}}
    \label{fig:assembly}\vspace*{-0.2cm}
\end{figure}

The gripper prototype is manufactured using 3D printing, excluding electronic components, screws, motors, tendon wires, and magnets. The exploded views of the gripper and finger assemblies are shown in \fig\ref{fig:assembly}, illustrating the modular design that allows separate finger assembly before attaching them to the palm. \rev{Considering the higher torque requirement by the thumb, a stronger motor with a torque rating of 10.6 Nm (DYNAMIXEL XM540-W270) is used as compared to 4.1 Nm (XM430-W350) in the other fingers.} The gripper has two-level control architecture (\fig\ref{fig:assembly}). \rev{At the lower level, a dedicated microcontroller unit (Arduino nano) reads joint sensors, while the DYNAMIXEL unit controls the motors with the help of motor encoders. At the master level, the gripper (housing two slave controllers) as a single unit is controlled by our gripper package inside Golem~\cite{golem_short} running on the PC.}

\begin{figure}[ht]
\begin{center}
\vspace*{0.1cm}
\begin{tabular}{cc}    
\begin{tabular}{c}
\hspace*{-0.4cm}\includegraphics[width=2.6cm]{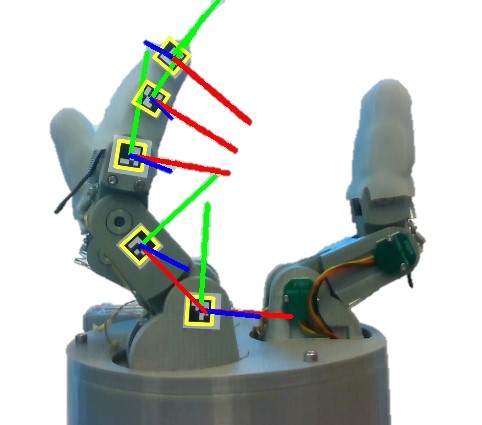}\vspace*{-0.3cm}
\\\hspace*{-0.6cm}\begin{minipage}{0cm}\subcaption{}\label{fig:sensor_model_a}\end{minipage}\\   
\hspace*{-0.4cm}\includegraphics[width=2.6cm]{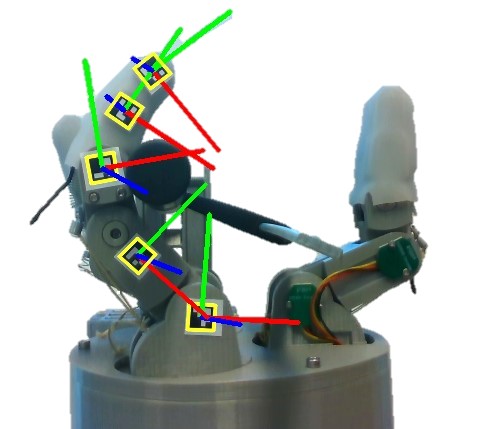}\vspace*{-0.3cm}
\\\hspace*{-0.6cm}\begin{minipage}{0cm}\subcaption{}\label{fig:sensor_model_b}\end{minipage}
\end{tabular}&

\begin{tabular}{c}
\hspace*{-0.8cm}\includegraphics[width=6.1cm]{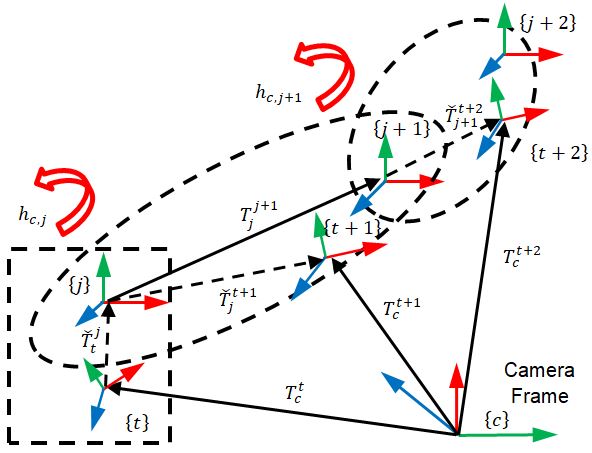}
\\\begin{minipage}{0cm}\subcaption{}\label{fig:sensor_model_c}\end{minipage}   
\end{tabular}

\end{tabular}   
\end{center}

\vspace*{-0.3cm}\caption{\rev{Sensor model estimation without (a) and with (b) obstacle, and kinematic relation between the frames (c).}}
\label{fig:sensor_model}\vspace*{-0.5cm}
\end{figure}
\textit{Sensor Model Estimation:}
\rev{Because grasp demonstrations require joint positions, the gripper is equipped with magnetic joint position sensors on the first two proximal joints.} 
%To demonstrate grasps, the gripper is equipped with magnetic joint position sensors on the first two proximal joints. 
The remaining joints' positions are estimated using a sensor model with experimentally determined parameters (\fig\ref{fig:sensor_model}). For $j-$th joint in $m-$th finger, the joint position $h_{c,*}$ and tendon displacement $l_{*}$ are modelled linearly using gain and offset parameters $\alpha_{*}, \zeta_{*} \in \mathbb{R}$:
\begin{equation}\label{eq:calib_model} \\
    h_{c, mj} = \alpha_{mj}\bar{s}_{mj} + \zeta_{mj}, \hspace*{0.4cm}\\
    l_{m} = \alpha_{m}\bar{r}_{m} + \zeta_{m} 
\end{equation}
% \begin{eqnarray}
% h_{c, mj} & = & \alpha_{mj}\bar{s}_{mj} + \zeta_{mj} \label{eq:calib1} \\
% l_{m} & = & \alpha_{m}\bar{r}_{m} + \zeta_{m} 
% \end{eqnarray}
where $\bar{s}_{*}$, $\bar{r}_{*}$ are sensor and motor position readings after applying low pass filter.

The $j-$th joint and $t-$th tracker $SE(3)$ frames are indexed by \{$j$\} and \{$t$\} as shown in \rev{\fig\ref{fig:sensor_model_c}}  \rev{(where $SE(3)$ denotes the special Euclidean group in 3D)}. Now, the tracker pose $T_{c}^t$ in camera frame is computed while the joint-to-joint transformation $T_{j}^{j+1}$ is formulated using the known gripper kinematics and \eq\eqref{eq:calib_model}. The unknown joint-to-tracker transformation $\Tilde{T}_{j}^t$ and the model parameters are estimated iteratively by minimising the least square error over the data samples:
\rev{
\begin{equation}
\alpha_{*}, \zeta_{*}, \Tilde{T}_{j}^* = \argmin_{(\alpha_{*}, \zeta_{*}, \Tilde{T}_{j}^*)}   \| T_c^{t}\Tilde{T}_{t}^j\prod_{i=0}^{j-1}T_i^{i+1}\Tilde{T}_{j}^{t+1} - T_c^{t+1}\| ^2
\end{equation}}
The model parameters enable estimating the joint positions of the last joints using motor displacement and sensor data.

%\subsection{Gripper sensors}
%
%\begin{figure}[ht]
%\begin{center}
%\begin{tabular}{cc}
%
%\hspace*{-0.6cm}\begin{tabular}{c}
%\begin{tabular}{c}
%\includegraphics[height=3.1cm,trim=0.0cm 0.0cm 0.0cm 0.0cm,clip]{resources/sensor_model/detected_pose_01/image_4_2.jpg}
%\end{tabular}
%\\\begin{minipage}{0cm}\subcaption{}\label{fig:sensor_model_a}\end{minipage}
%\end{tabular}
%
%\hspace*{-0.8cm}\begin{tabular}{c}
%\begin{tabular}{c}
%\includegraphics[height=3.1cm,trim=0.0cm 0.0cm 0.0cm 0.0cm,clip]{resources/sensor_model/detected_pose_02/image_2_2.jpg}
%\end{tabular}
%\\\begin{minipage}{0cm}\subcaption{}\label{fig:sensor_model_b}\end{minipage}
%\end{tabular}
%
%\end{tabular}
%\end{center}
%
%\caption{Sensor model parameters' estimation without (a) and with (b) obstacle.}
%\label{fig:sensor_model}
%\end{figure}
%
%In this work, we attempt to demonstrate grasps on a real robot to create contact models. For this purpose, we equipped our gripper with magnetic joint position sensors, but to minimise finger stiffness, only on the first two proximal joints. The remaining joints have no sensors, but their positions are determined based on a \textit{sensor model} with parameters estimated experimentally (see \fig\ref{fig:sensor_model}).

\section{Grasp learning with underactuated gripper}

In~\cite{kopicki2019learning}, we proposed an approach for dexterous grasping that can generate grasps for novel objects from a single camera view and from even one example. While this approach works well with position controlled robots, it cannot be directly applied to tendon-driven underactuated grippers. \revv{In the following work we show how to overcome this limitation. }
%follwoing, how this work overcomes this limitation. 
\sect\ref{sec:contact_models} briefly describes the dexterous grasp learning algorithm that was originally proposed in~\cite{kopicki2019learning}. \sect\ref{sec:contact_optimisation} introduces our contact optimisation method to learn more accurate and reliable contact models. Finally, \sect\ref{sec:reconfiguration_planning} presents the trajectory planner, which sets the \rp joints to the desired configuration.

\subsection{Dexterous grasp learning with contact models}\label{sec:contact_models}

Our approach relies on learning \textit{contact models} that represent robot links - \textit{ training object} local spatial relations. Then, given a novel \textit{test object}, each contact model is transformed into a \textit{query density} which represents the poses of a given robot link. Grasps are generated by finding maxima of \textit{products} of suitable query densities given robot kinematics.

%Contact models are learned only within the neighbourhood of a particular robot link $i$ with 3D pose $s_i \in SE(3)$  (see \fig\ref{fig:contact_recfield} solid yellow block), called \textit{contact receptive field} $\rf_i$ (depicted as a dashed line around the link). The black dots are \textit{surface features} $x$ from a point cloud representing the object's surface:
\rev{Contact models are learned only within the neighbourhood of a particular robot link $i$ with 3D pose $s_i \in SE(3)$, called \textit{contact receptive field} $\rf_i$ (see \fig 6 and \eq 10-11 in \cite{kopicki2019learning}). The object's surface is represented by a point cloud - a set of \textit{surface features} $x$, where each $x$ consists of SE(3) frame $v$ with position $p$ and orientation $q$, and a surface descriptor vector $r\in \mathbb R^2$ which is assumed here to be the local principal curvature.}
%\begin{subequations}
%\begin{align}
%x \equiv (v, r) \textnormal{ where } v\in SE(3), r\in \mathbb R^2 \label{eq:surface_feature}\\
%v \equiv (p, q) \textnormal{ where } p\in \mathbb R^3, q\in SO(3) \label{eq:surface_pose}
%\end{align}
%\end{subequations}
%so, each surface feature consists of a 3D frame $v$ with position $p$ and orientation $q$, and a surface descriptor vector $r$ which is assumed here to be the local principal curvature.
%Formally, the contact receptive field $\rf_i$ is a function of surface feature pose $v$:
%\begin{equation}
%\rf_{i} : SE(3) \rightarrow [0, 1]
%\label{eq:contact_recfield_model}
%\end{equation}
%\noindent the value of which determines the relevance of a particular surface feature $x = (v, r) = ((p, q), r)$ to a given gripper-link $i$ in terms of the likelihood of physical contact. Specifically:
%\begin{equation}
%\rf_i(v|\lambda_i,\delta_i) = \begin{cases}\exp(-\lambda_i d_{\rf_i}(p)^2) \quad &\textnormal{ if } d_{\rf_i}(p) < \delta_i\\
%0 \quad &\textnormal{ otherwise},\end{cases}
%\label{eq:contact_recfield_func}
%\end{equation}
%where $\delta_i > 0$ is a receptive field diameter, $\lambda_i > 0$ and $d_{\rf_i}(p)$ is a function that measures the distance of $p$ to $a_i$ - a point on the surface of link $i$ that is closest to $p$. In this way, the contact receptive field will only take account the local shape, while falling off smoothly.
%
%\begin{figure}[t]
%\centering\includegraphics[width=4cm]{resources/contact}
%\caption[Contact receptive field]{Contact receptive field.}
%\label{fig:contact_recfield}\vspace*{-0.4cm}
%\end{figure}
All \textit{point clouds} are captured from a single camera view and approximated as kernel densities over features $x$. For a point cloud used for training of a particular grasp $g$:
%The set of features visible from the $m$\textsuperscript{th} view define a joint probability distribution as follows
\begin{equation}\label{eq:object_pdf}
V^g(x)\approx\sum_{j=1}^{N_{g}}w_{j}^g\mathcal{K}(x\mid x_{j}^g,\sigma)
\end{equation}
%\begin{equation}\label{eq:object_density}
%\pdf(x) \simeq \sum_{j=1}^N w_j \mathcal{K}(x| x_{j}, \sigma),
%\end{equation}
%$\pdf(x) \simeq \sum_{j=i}^{K} w_j \mathcal{K}(x|x_j , \sigma)$,
where $\sigma \in \mathbb{R}^3$ is the kernel bandwidth, $N_{g}$ is a number of features in the point cloud , and $w_j^g \in \mathbb{R}^+$ is a weight associated to $x_j^g$ such that $\sum_{j}w_j^g = 1$. Kernel $\mathcal{K}$ is factorized into three functions to cope with the separation of $x$ into position $p$, a quaternion $q$ for orientation, and a surface descriptor $r$.
Let us denote the kernel mean point $\mu =(\mu_p , \mu_q , \mu_r )$, and the kernel bandwidth $\sigma = (\sigma_p, \sigma_q, \sigma_r)$, decomposed into the three factors. The kernel is defined as
\begin{equation}\label{eq:factorised_kernel}
\mathcal{K}(x|\mu, \sigma) = \mathcal{N}_ 3 (p|\mu_p , \sigma_p )\Theta(q|\mu_q , \sigma_q )\mathcal{N}_2 (r|\mu_r , \sigma_r )
\end{equation}
where $\mathcal{N}_n$ is an $n$-variate isotropic Gaussian kernel, and $\Theta$ corresponds to a pair of antipodal von Mises-Fisher distributions which form a Gaussian-like distribution on $SO(3)$ %(\cite{fisher1953dispersion}\rev{ and where $SO(3)$ denotes the special orthogonal group in 3D}).
(\revv{$SO(3)$ denotes the special orthogonal group in 3D, see \cite{fisher1953dispersion}})
% (for details see \cite{fisher1953dispersion}). 

Let us denote by $u_{ij} = (p_{ij}, q_{ij})$ the pose of link $i$ relative to the pose $v_j$ of the $j$\textsuperscript{th} surface feature $x_j$. $u_{ij}$ can be found from the relation
%\begin{equation}
%u_{ij} = v_j^{-1} \circ s_i,
%\end{equation} 
$u_{ij} = v_j^{-1} \circ s_i$, where $s_i$ is the pose of link $i$ in the world frame, $\circ$ is the pose composition operator, and $v_j^{-1}$ is the inverse of $v_j$. %(\eq 12 in \cite{kopicki2019learning}).
For a particular grasp $g$, $i$\textsuperscript{th} link and point cloud \eq\eqref{eq:object_pdf}, the \textit{contact model} $M_{i}^g$ for  is computed as
\begin{equation}\label{eq:contact_model}
M_{i}^g(u,r)= \frac{1}{Z}\sum_{j=1}^{N_{g}} w_{j}^g F_i(v_{j}^g)\mathcal{K}(u,r|u_{ij}^g,r_{j}^g,\sigma)
\end{equation}
with a normalizing constant $Z>0$, $u=(p,q)$, $\mathcal{K}$ defined in \eq{\ref{eq:factorised_kernel}}, and $F_i$ is a contact receptive field for $i$\textsuperscript{th} link. 

%$\rf$ associated with the $i^{th}$ hand-link $\rl_i$ (solid yellow block) with link pose $s_i$. The black dots are samples from the surface of an object. The distance $a$ between feature $v$ and the closest point $a$ on the link's surface is shown. The rounded rectangle illustrates the cut-off distance $\delta_i$. The poses $v$ and $s_i$ are expressed in the world frame $W$. The arrow $u$ is the pose of $\rl_i$ relative to the frame for the surface feature $v$.

%\begin{figure}[ht]
%\begin{center}
%\begin{tabular}{cc}
%
%\begin{tabular}{c}
%\begin{tabular}{c}
%\includegraphics[height=4.5cm,trim=0.0cm 0.0cm 0.0cm 0.0cm,clip]{resources/contact_model.jpg}
%\end{tabular}\\
%%\hspace*{0.5cm}\begin{minipage}{0cm}\subcaption{}\label{fig:contact_model_a}\end{minipage}
%%\hspace*{2.8cm}\begin{minipage}{0cm}\subcaption{}\label{fig:contact_model_b}\end{minipage}
%%\hspace*{2.0cm}\begin{minipage}{0cm}\subcaption{}\label{fig:contact_model_c}\end{minipage}
%\end{tabular}
%
%\end{tabular}
%\end{center}\vspace*{-0.2cm}
%\caption{Contact models.}\label{fig:contact_model}
%\vspace*{-0.4cm}
%\end{figure}

\rev{For a given robot link $i$, a demonstrated grasp $g$, a corresponding contact model $M_{i}^g$, and a test point cloud that describes a novel scene, the \textit{query density} represents possible 3D poses $s$ (in world frame) of the $i$\textsuperscript{th} robot's link relative to the cloud. Intuitively, the robot link is ``encouraged'' to make contact in places with surface properties similar to those observed during training (see \fig 7 in \cite{kopicki2016learning}).}
%In the example shown in \fig\ref{fig:contact_model}, a contact model trained on a link touching training object parts with greater curvature, will result in a query density with higher values around more ``curvy'' parts of the test object (sampled link's positions in red).% The query density is used both to generate and evaluate the likelihood of a candidate grasp.

We implement the query density as convolving two densities: contact model density \eq\eqref{eq:contact_model} and the \textit{object model density} computed from the test point cloud \eq\eqref{eq:object_pdf} (details in~\cite{kopicki2019learning}). We approximate the query density as a kernel density estimation with $K_Q$ centred on the set of weighted link poses sampled from the contact model density and the object density
\begin{equation}\label{eq_query_densities}
Q_{i}^g(s)\approx\sum_{j=1}^{K_Q}w_{ij}N_3(p\mid \hat{p}_{ij},\sigma_p)\Theta(q\mid \hat{q}_{ij},\sigma_q)
\end{equation}
with $j$\textsuperscript{th} kernel centre $(\hat{p}_{ij},\hat{q}_{ij})=\hat{s}_{ij}$, sampled as 
%\begin{equation}\label{eq_query_sampling}
%(\hat{v}_{j}, \hat{r}_{j})\sim V^g(v,r),\; \hat{u}_{ij}\sim M_{i}^g(u\mid\hat{r}_j),\; w_{ij}=M_{i}^g(\hat{r}_j)
%\end{equation}
$(\hat{v}_{j}, \hat{r}_{j})\sim V^g(v,r)$, $\hat{u}_{ij}\sim M_{i}^g(u\mid\hat{r}_j)$ with $w_{ij}=M_{i}^g(\hat{r}_j)$.
%\begin{subequations}
%\begin{align}
%(\hat{v}_{j}, \hat{r}_{j})&\sim V^g(v,r)\\
%\hat{u}_{ij}&\sim M_{i}^g(u\mid\hat{r}_j),\\
%w_{ij}&=M_{i}^g(\hat{r}_j).
%\end{align}
%\end{subequations}

The \textit{gripper configuration model} $C^g$ is an evaluative-generative model of possible joint configurations of the gripper for grasp $g$. $C^g$ is formulated as a Gaussian-like density around configuration of the demonstrated grip pose $g$. Let us also denote $h=(h_w,h_c)$, where $h_w\in SE(3)$ is the wrist pose and $h_c\in \mathbb{R}^{D^h}$ is the gripper joint configuration ($D^h=14$). 

The initial grasp is generated first by sampling the grasp type from a set of $N_G$ trained grasps $g^*\sim {1..N_G}$, then sampling query density $i^*$ that belongs to the set of densities for grasp $g^*$: $Q_{i^*}^{g^*}\sim \mathcal{Q}^{g^*}$. This determines the pose of a robot link that corresponds to $Q_{i^*}^{g^*}$. $h_c^*$ is obtained by sampling the configuration model $h_c^*\sim C^{g^*}$, this enables computing poses of the all remaining links together with the wrist pose $h_w^*$.

The above procedure repeated several times allows to generate multiple independent grasps. Each grasp is then optimised by maximising the product of the likelihood of the query densities and the hand configuration density:
\begin{equation}\label{eq:grasp_selection}
h^*=\argmax_{(h_w,h_c)}C^g(h_c)\coll(h_w, h_c)\prod_{Q_{i}^g\in\mathcal{Q}^{g}}Q_{i}^g(\kin_i(h_w,h_c))
\end{equation}
%where $C^g(h)$ is the gripper configuration model, $Q_{i}^g$ are the query densities from $\mathcal{Q}^{g}$ for that specific object-grasp $g$.
where $\coll(h_w, h_c)$ is the \textit{collision model} that penalises gripper-point cloud collisions in a soft manner:
\begin{subequations}
\begin{align}
\label{eq:collision_model_a}
\coll(h_w, h_c) &= \prod_{i=1..D^h}\exp(-\gamma \coll_i(h_w, h_c))\\
\label{eq:collision_model_b}
\coll_i(h_w, h_c) &= \sum_{j=1..N}\left(\exp(\beta d_{\coll_i}(p_j, h_w, h_c)^2) - 1 \right)
\end{align}
\end{subequations}
where $\gamma, \beta > 0$, $d_{\coll_i}(p, h_w, h_c)^2)$ is the penetration depth of link $i$ at pose $(h_w, h_c)$ by point $p$, and where $p_j$ are points of the test point cloud of size $N$.

\subsection{Kinaesthetic contact optimisation}\label{sec:contact_optimisation}
While this approach works well with position controlled robots, it cannot be directly applied to tendon-driven underactuated grippers. This is because the motor command - gripper posture relation can only be estimated experimentally and depends on many additional factors, such as the properties of the manipulated object.
Real demonstrations allow the robot to learn motor commands and stable grips. However, the robot's links poses were not always accurately estimated, mainly due to sensor model limitations and imperfect camera calibration. As a result, some contact models were not created because the target object was too far away, or there were created, but colliding with the object (\fig\ref{fig:training_b} training example 6 and 5).

\rev{The k}\textit{inaesthetic contact optimisation} procedure adjust\rev{s} gripper joint configuration $h_c$ and its 3D pose $h_w$ \rev{w.r.t.} the training object \eq\eqref{eq:object_pdf}, minimising collisions while maximising link-object contacts 
% , but only locally, in some $\epsilon$ neighbourhood of a given grasp training example configuration $h^g \equiv (h_w^g,h_c^g)$:
locally, in $\epsilon$ neighbourhood of a training grasp example $h^g \equiv (h_w^g,h_c^g)$:
\begin{subequations}
\begin{align}
\label{eq:contact_optimisation_a}
&h^{g*} = \argmax_{h \in (h^g - \epsilon, h^g + \epsilon)}(\mathcal{M}^g(h_w, h_c) - \zeta \sum_{i=1..D^h} \coll_i(h_w, h_c))\\
\label{eq:contact_optimisation_b}
&\mathcal{M}^g(h_w, h_c) = \sum_{i=1..D^h}\sum_{j=1..N^g} w_j^g \rf_i(v^g_j|h_w, h_c)
\end{align}
\end{subequations}
where $\mathcal{M}$ is a \textit{contact mass} function \eq\eqref{eq:contact_optimisation_b}, $\coll_i$ is a collision model \eq\ref{eq:collision_model_b} with receptive field $\rf_i$ and an explicit dependency on $(h_w, h_c)$ shown.

\subsection{Planning for \rp joint re-configuration}\label{sec:reconfiguration_planning}

While \rp joints significantly improve gripper dexterity, by design they require external forces to reconfigure. Those forces can be generated by the robot itself by pushing its fingers against an obstacle as shown in \fig\ref{fig:reconfig}.

%In the robot configuration space, we defined the \textit{reconfiguration trajectory}, the execution of which leads to the expected changes in the gripper configuration, as a linear function of the progress of the trajectory. Here, each passive joint requires two such trajectories - increasing and decreasing joint rotation. Furthermore, we created a \textit{planner} that finds the shortest path that results in the desired configuration of all passive joints, including empty paths if no changes are required.

In the arm configuration space $\mathbb{R}^{D^a}$, we defined the \textit{reconfiguration trajectory} $(a_{ci}^R)_{i=1..N^R}$ as an ordered sequence of $N^R$ waypoints. We assume that the execution of the entire trajectory changes the gripper configuration to $h_c^{end}$, as long as the initial gripper configuration $h_c^{init}$ is in some predefined range $(h_c^{begin}, h_c^{end})$. This models the effect of robot moving through free space without changing $h_c^{init}$ and then pushing against the obstacle until $h_c^{init}$ changes to $h_c^{end}$. We also assume that the change of $h_c^{init}$ with respect to $(h_c^{begin}$ linearly depends on the changes of an Euclidean distance in the arm configuration space $\mathbb{R}^{D^a}$ along trajectory $(a_{ci}^R)$. We can then create a linear \textit{trajectory interpolator} that given a desired gripper configuration $h_c^*$ finds waypoint $a_{c}^*$ that lies somewhere in between waypoints $(a_{ci}^R)$:
\begin{equation}\label{eq:reconfig_interpol}
f^{int}(h_c^*|h_c^{begin}, (a_{ci}^R)) \rightarrow a_{c}^* 
\end{equation}
such that $f^{int}(h_c^{begin}) = a_{c1}^R$ and $f^{int}(h_c^{end}) = a_{cN^R}^R$. In this way, if the target gripper configuration $h_c^{*}$ is different than $h_c^{end}$, then one can find $a_c^* = f^{int}(h_c^*)$, and set it as the final waypoint on the trajectory that leads to $h_c^*$. 

To avoid online planning with nearby obstacles, we define the \textit{approach trajectory} $(a_{ci}^A)$ that leads to the first waypoint $a_{c1}^R$ of the reconfiguration trajectory, assuring a collision-free path. Let's define a tuple that collects both the trajectories and the range $(h_c^{begin}, h_c^{end})$, we will refer to it as a trajectory:
\rev{
\begin{multline}
\label{eq:reconfig}
\mathbf{T}=\left\{h_c^{begin}, h_c^{end}, (a_{ci}^R: a_{ci}^R \in \mathbb{R}^{D^a})_{i=1..N^R},\right.\\
\left.(a_{ci}^A: a_{cN^A}^A=a_{c1}^R, a_{ci}^A \in \mathbb{R}^{D^a})_{i=1..N^A}\right\}\textnormal{.}
\end{multline}}
Due to the gripper design, each \rp joint requires two such trajectories - an increasing and decreasing joint rotation, i.e., four trajectories to reconfigure independently two \rp joints.

Given the target gripper configuration $h_c^*$, at most two (out of four) trajectories will be selected to execute, since a single \rp joint position is either lower, higher, or equal (with some tolerance $\delta > 0$) to the target one. 
If a specific trajectory $\mathbf{T}_k$ is selected, then it is also updated with \eq\ref{eq:reconfig_interpol} with the new final waypoint for the arm movement. For each trajectory, a \textit{rollback trajectory} is also created by simply using the waypoints in the reverse order. Moreover, approach trajectories for different $\mathbf{T}_{\{k, l : k \neq l\}}$ can also be shortened if they share the same waypoints. The resulting trajectories are then executed sequentially in an open loop, as in the example from \fig\ref{fig:reconfig} .

\section{Results}

\subsection{Robotic gripper}
Gripper experiments have been carried out to validate its capabilities in real-world grasping scenarios. Fig.~\ref{fig:locking} shows the locking of \rp joints. Initially, The joints can be moved by applying a small external force (\fig\ref{fig:locking} top). Once the tendons start to close the fingers, the fingers can not be moved as the locking mechanism locks the joints(\fig\ref{fig:locking} bottom).

%\begin{figure}[ht]
%	
%	\begin{center}
%		\begin{tabular}{cc}
%			\begin{tabular}{cc}%trim=left botm right top
%				\begin{tabular}{c}
%					\hspace*{-0.6cm}\includegraphics[height=2.0cm, width=2.0cm]{resources/gripper_results/Locking_a.jpg}\\\hspace*{-0.0cm}
%				\end{tabular}&        
%				
%				\begin{tabular}{c}
%					\hspace*{-0.8cm}\includegraphics[height=2.0cm,width=2.0cm]{resources/gripper_results/Locking_b.jpg}\\\hspace*{-0.0cm}
%				\end{tabular}\vspace*{-0.7cm}\\
%				\begin{tabular}{c}
%					\hspace*{1.0cm}\begin{minipage}{0cm}\subcaption{}\label{fig:locking_a}\end{minipage}
%				\end{tabular}
%			\end{tabular}& 
%			\begin{tabular}{cc}%trim=left botm right top
%				\begin{tabular}{c}
%					\hspace*{-0.8cm}\includegraphics[height=2.0cm, width=2.0cm]{resources/gripper_results/Locking_c.jpg}\\\hspace*{-0.0cm}
%				\end{tabular}&        
%				
%				\begin{tabular}{c}
%					\hspace*{-0.8cm}\includegraphics[height=2.0cm, width=2.0cm]{resources/gripper_results/Locking_d.jpg}\\\hspace*{-0.0cm}
%				\end{tabular}\vspace*{-0.7cm}\\
%				\begin{tabular}{c}
%					\hspace*{1.0cm}\begin{minipage}{0cm}\subcaption{}\label{fig:locking_b}\end{minipage}
%				\end{tabular}
%			\end{tabular}    
%			
%		\end{tabular}
%	\end{center}\vspace*{-0.4cm}
%	
%	\caption{Initially \rp joints free to move (a) tendon locked the joints (b). }
%	\label{fig:locking}\vspace*{-0.4cm}
%\end{figure}

\begin{figure}[ht]
	\begin{center}\vspace*{-0.1cm}
		\begin{tabular}{c}
			\begin{tabular}{cc}
				\begin{tabular}{c}
					\hspace*{-0.6cm}\includegraphics[height=1.7cm, width=2.0cm]{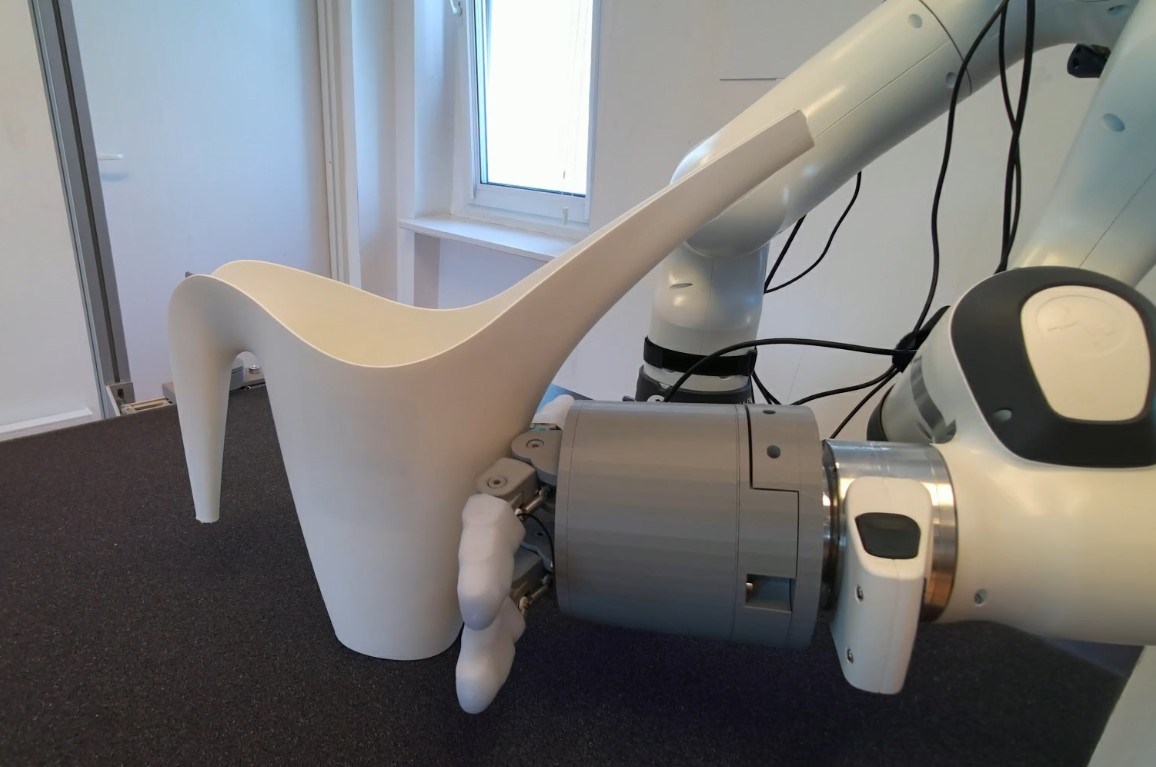}\\\hspace*{-0.0cm}
				\end{tabular}&
				
				\begin{tabular}{c}
					\hspace*{-0.8cm}\includegraphics[height=1.7cm, width=2.0cm]{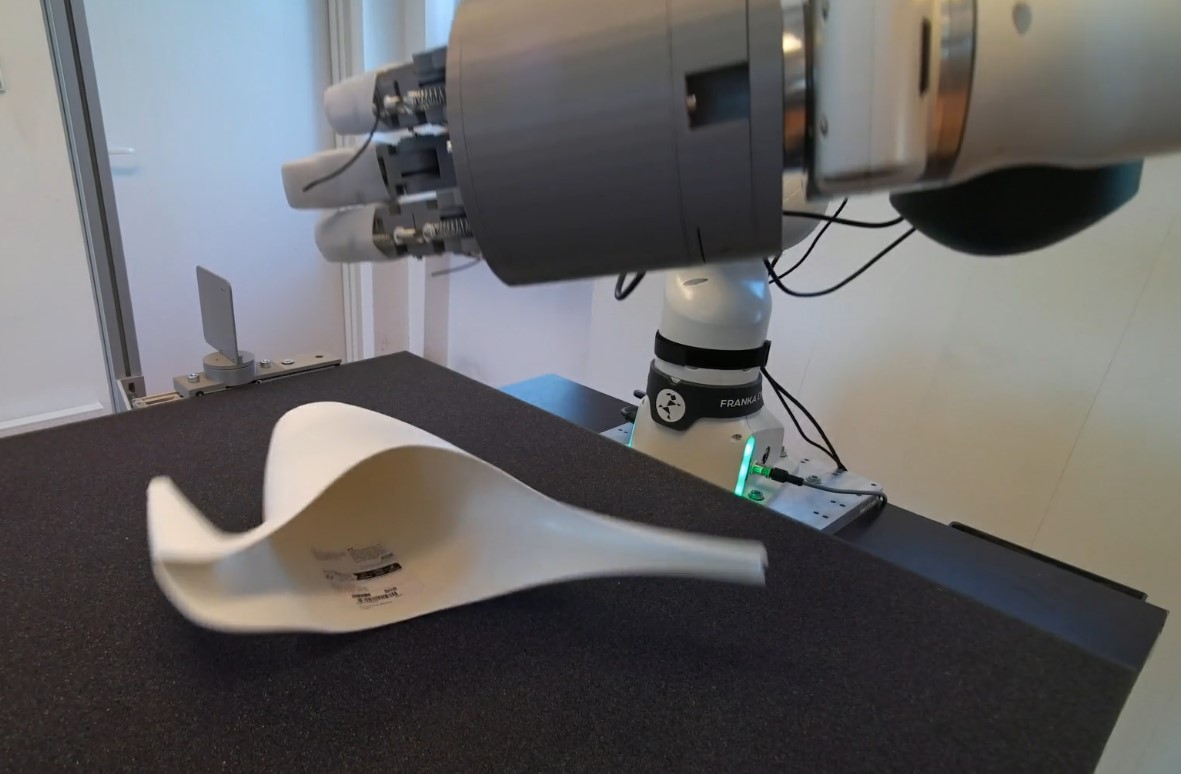}\\\hspace*{-0.0cm}
				\end{tabular}
                \end{tabular}
                \begin{tabular}{cc}
				
				\begin{tabular}{c}
					\hspace*{-0.4cm}\includegraphics[height=1.7cm, width=2.0cm]{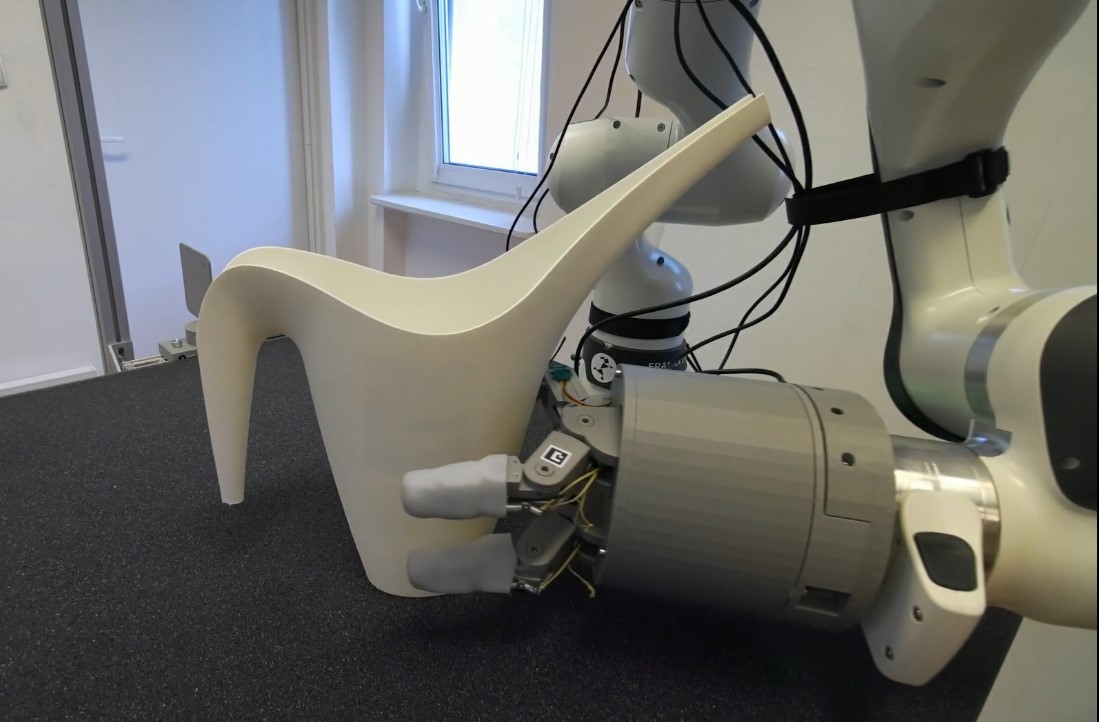}\\\hspace*{-0.0cm}
				\end{tabular}&
				
				\begin{tabular}{c}
					\hspace*{-0.8cm}\includegraphics[height=1.7cm, width=2.0cm]{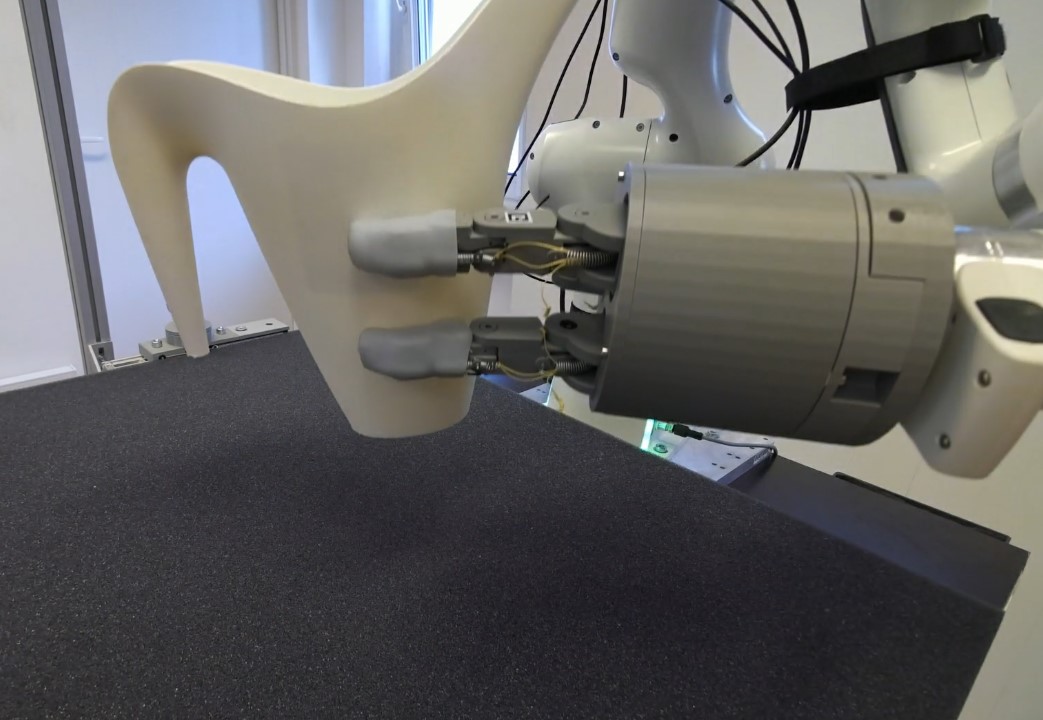}\\\hspace*{-0.0cm}
				\end{tabular}
			\end{tabular}
			
		\end{tabular}
	\end{center}\vspace*{-0.8cm}
	
	\caption{Grasp failure due to outward force by fingers while fingers closing not parallel (left) Successful grasp while fingers closing parallel (right). }
	\label{fig:parallel}\vspace*{-0.2cm}
\end{figure}
\begin{figure}[ht]
	\begin{center}\vspace*{-0.2cm}
		\begin{tabular}{c}
			\begin{tabular}{cc}
   %              \begin{tabular}{c}
			% 	\hspace*{-0.6cm}\begin{minipage}{0cm}\subcaption{}\label{fig:collision_b}\end{minipage}
			% \end{tabular}&
   
				\begin{tabular}{c}
					\hspace*{-0.6cm}\includegraphics[height=1.7cm, width=2.0cm]{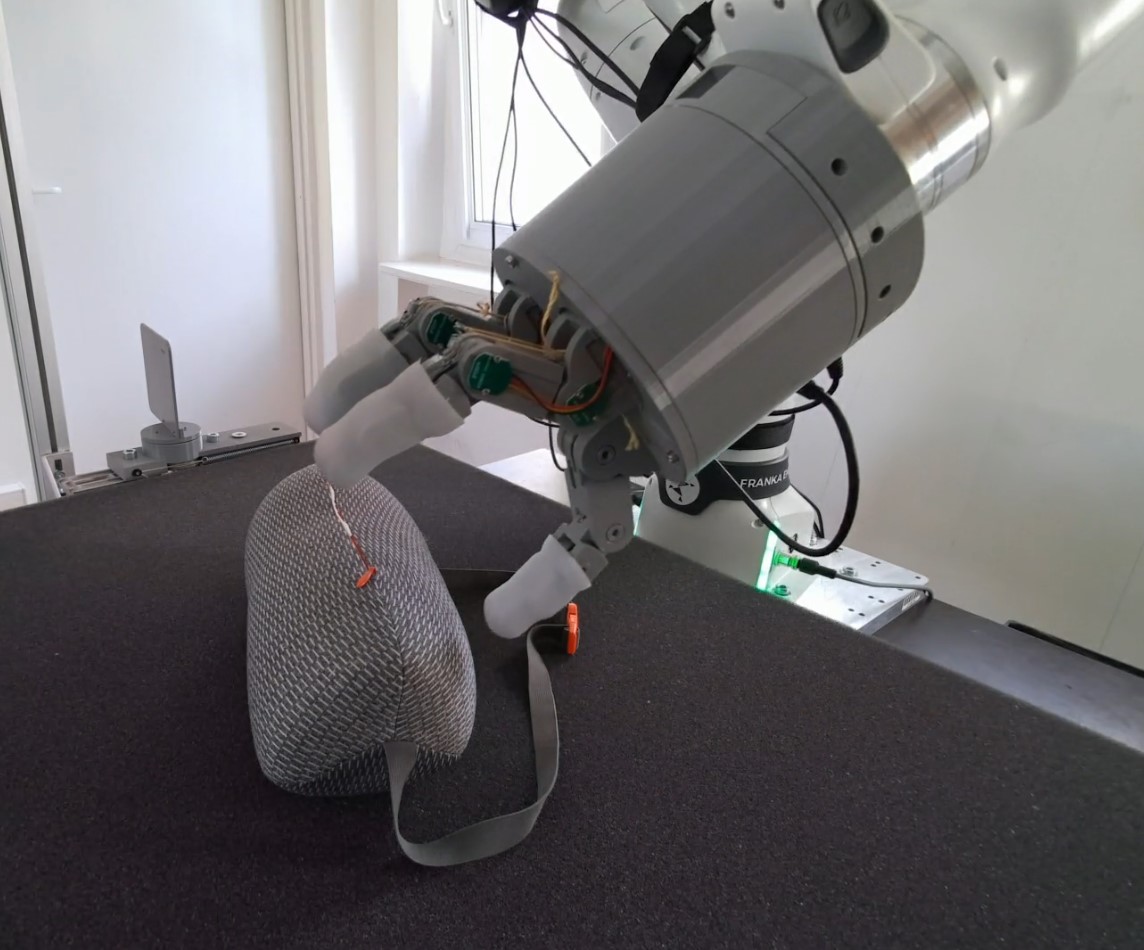}\\\hspace*{-0.0cm}
				\end{tabular}&
				
				\begin{tabular}{c}
					\hspace*{-0.8cm}\includegraphics[height=1.7cm, width=2.0cm]{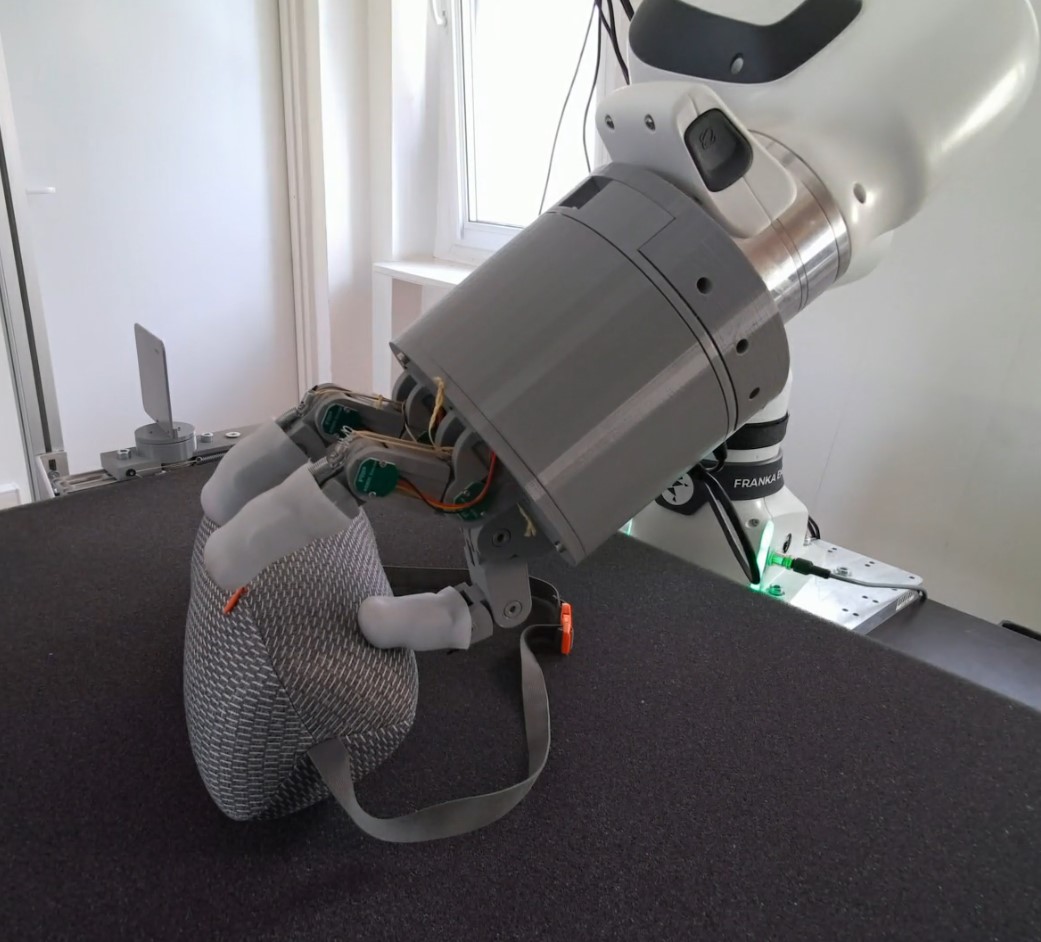}\\\hspace*{-0.0cm}
				\end{tabular}
                \end{tabular}
                \begin{tabular}{cc}
			% 	\begin{tabular}{c}
			% 	\hspace*{-1.6cm}\begin{minipage}{0cm}\subcaption{}\label{fig:collision_b}\end{minipage}
			% \end{tabular}&
   
				\begin{tabular}{c}
				\hspace*{-0.4cm}\includegraphics[height=1.7cm, width=2.0cm]{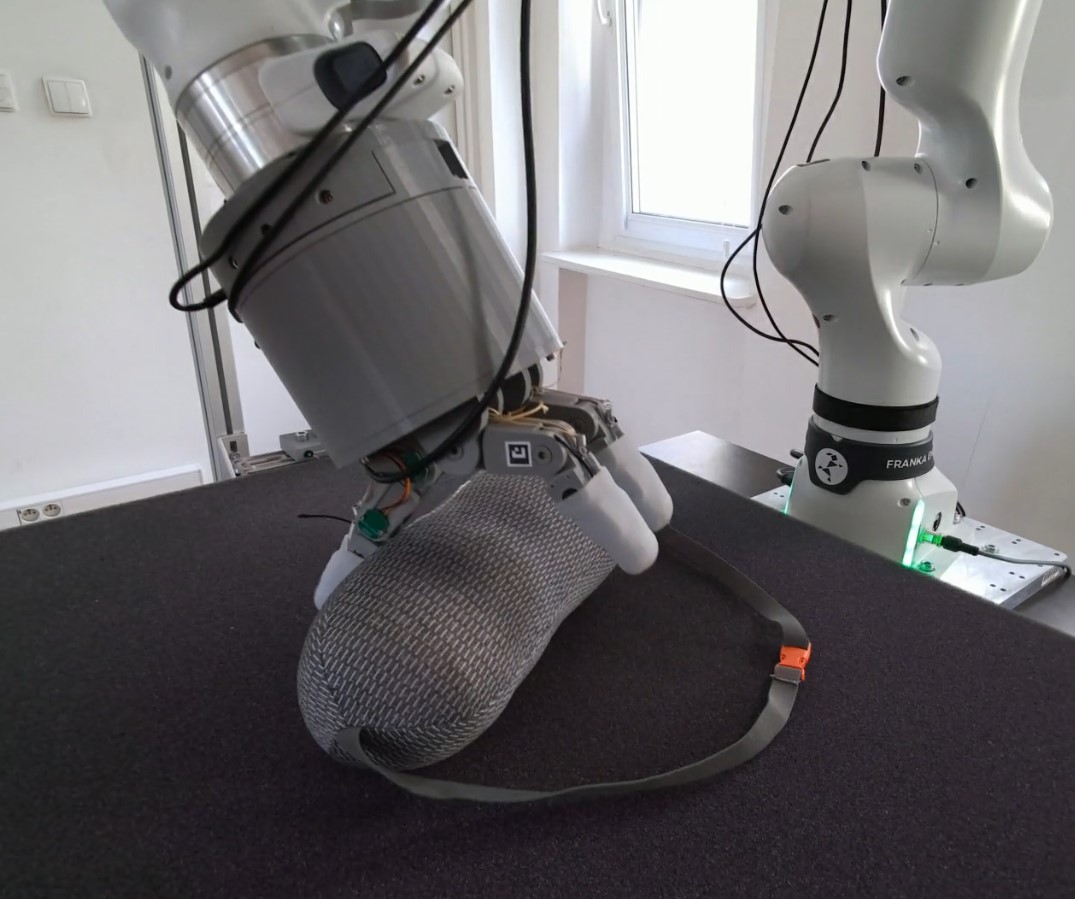}\\\hspace*{-0.0cm}
				\end{tabular}&
				
				\begin{tabular}{c}
					\hspace*{-0.8cm}\includegraphics[height=1.7cm, width=2.0cm]{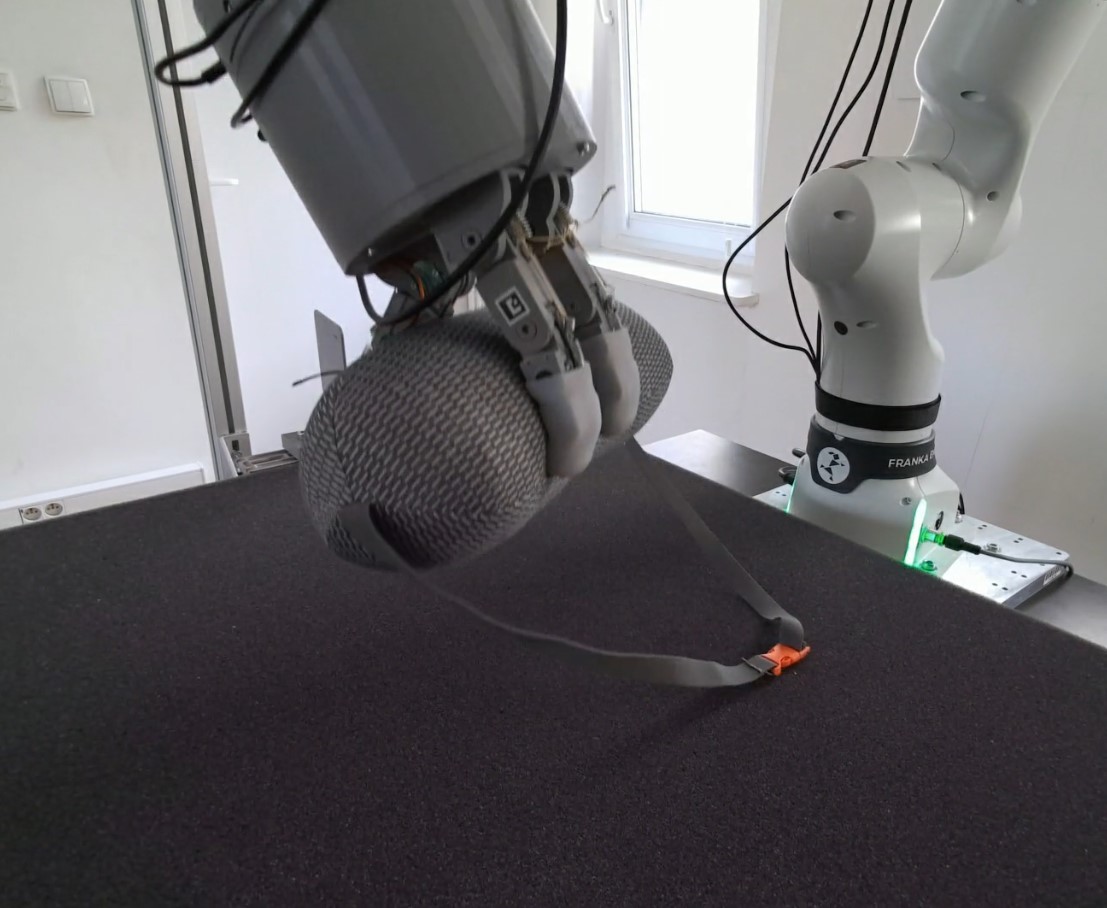}\\\hspace*{-0.0cm}
				\end{tabular}
			\end{tabular}
			
		\end{tabular}
	\end{center}\vspace*{-0.4cm}\vspace*{-0.4cm}
	
	\caption{Compliant joints helping fingers from collision damage while finger forward movement resulting in grasp failure (left) and backward movement in grasp success (right). }
	\label{fig:collision}\vspace*{-0.2cm}
\end{figure}
Handling lightweight objects, especially grasping from sideways, has always been one of the challenges faced by underactuated adaptive grippers. Most of the underactuated grippers are designed in such a way that they are good at enveloping grasp or finger-tip grasp. This is achieved by moving all the joints simultaneously for the former or moving one joint after another starting from the first joint for the latter. Although the former types of grippers do well with enveloping, they perform poorly on small objects or parallel grasps~\cite{Ko2020}. The latter types of grippers struggle with side grasp as fingers apply outward force on the object pushing it away while grasping as shown in \fig\ref{fig:parallel}(left). The tendon routing of the gripper in \fig\ref{fig:parallel}(left) is similar to the soft-gripper~\cite{hirose1978development} and without the four-bar mechanism in the second joint. The opposing forces from the fingers for parallel fingers help to successfully grasp the object as shown in \fig\ref{fig:parallel}(right).

Fig.~\ref{fig:collision} shows how \rp joint compliance in the second joint helps the fingers from collision damage. In the first case(\fig\ref{fig:collision} left), the impact causes forward joint movements preventing the fingers from securing the object. In the second case(\fig\ref{fig:collision} right), the finger joints move backwards allowing the fingers to secure the object resulting in a successful grasp.  Most importantly, the gripper safely performs both tasks without any damage from the impact.

\subsection{Robotic grasping experiments}

\begin{figure}[ht]
%\vspace*{-0.4cm}
\vspace*{0.1cm}
\begin{center}
\hspace*{-0.1cm}\begin{tabular}{cc}%trim=left botm right top
\includegraphics[height=4.0cm, trim=0.0cm 0.0cm 0.0cm 0.0cm,clip]{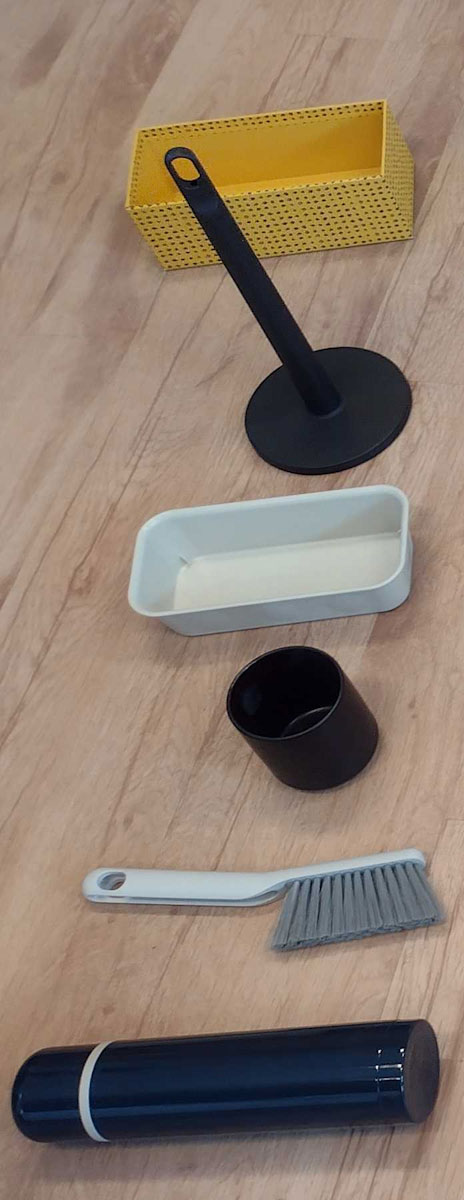}&
\hspace*{-0.2cm}\includegraphics[height=4.0cm, trim=0.0cm 0.0cm 0.0cm 0.0cm,clip]{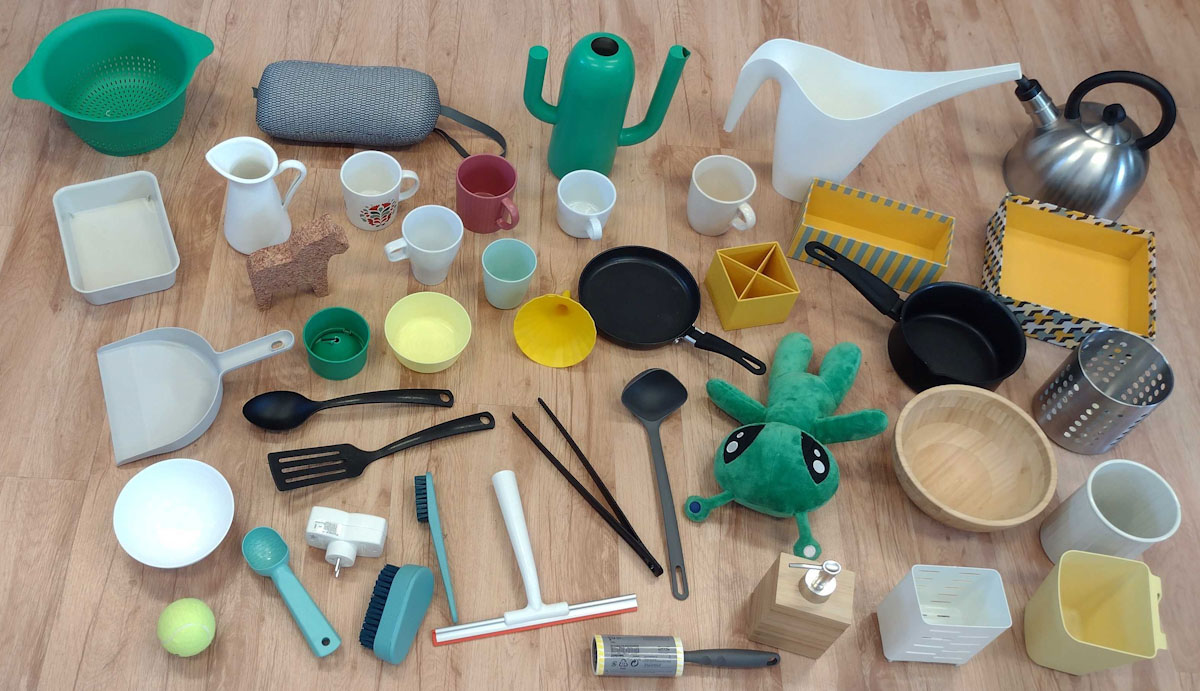}
\end{tabular}
\end{center}\vspace*{-0.4cm}

\caption{IKEA object sets used for training (left) and test.}
\label{fig:objects}\vspace*{-0.2cm}
\end{figure}

\subsubsection{Demonstrations}
The robot was demonstrated with 8 training trajectories, each labelled with number 1 to 8 (\fig\ref{fig:training_a}), all on 6 IKEA objects (\fig\ref{fig:objects}). Each trajectory is different in terms of the gripper configuration in the target grip, which includes rotational changes of the \rp joints, the local shape of the object, but also the motor commands that were used to control the gripper fingers. 

\subsubsection{Contact optimisation}
\begin{figure}[ht]
\begin{center}
\begin{tabular}{cc}
\vspace*{-0.2cm}
\begin{tabular}{c}
\includegraphics[width=6.0cm, trim=0.0cm 0.0cm 0.0cm 0.0cm,clip]{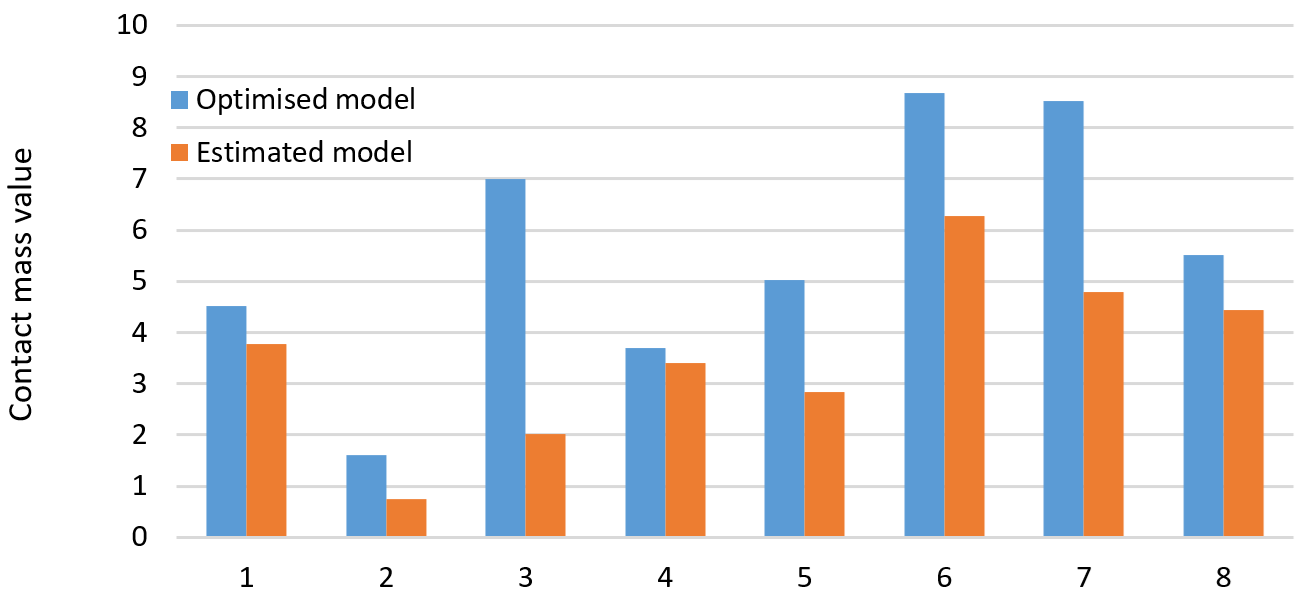}\\
\includegraphics[width=6.0cm, trim=0.0cm 0.0cm 0.0cm 0.0cm,clip]{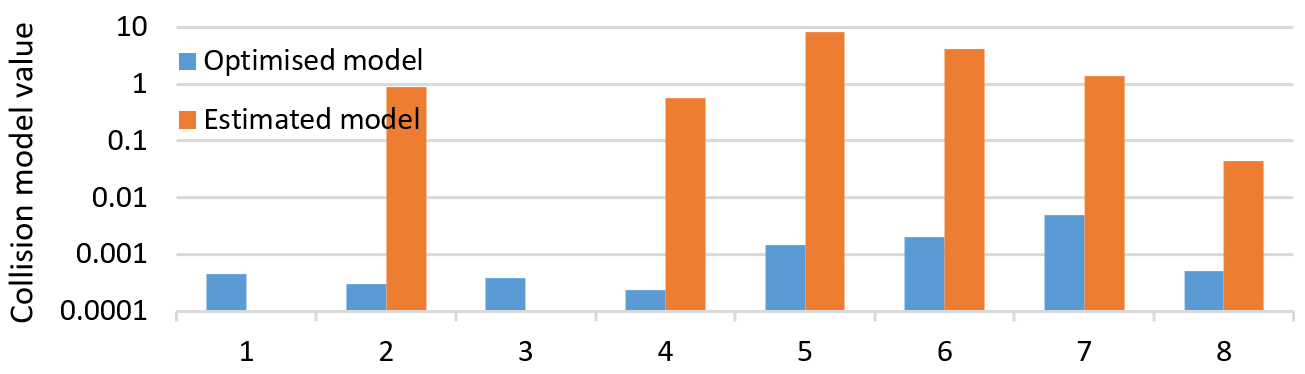}
\end{tabular}
&\vspace*{-0.2cm}\begin{tabular}{c}
\hspace*{-0.5cm}\vspace*{2cm}\begin{minipage}{0cm}\subcaption{}\label{fig:contact_optimisation_a}\end{minipage}\\
\hspace*{-0.5cm}\begin{minipage}{0cm}\subcaption{}\label{fig:contact_optimisation_b}\end{minipage}
\end{tabular}
\end{tabular}
\end{center}
\caption{\rev{Contact optimisation for grasp training examples.}}\label{fig:contact_optimisation}
\vspace*{-0.3cm}
\end{figure}

We trained two sets of contact models: \textit{estimated models} based entirely on the estimated gripper configuration with using sensor model (\sect\ref{sec::design}), and \textit{optimised models} based on kinaesthetic contact optimisation procedure (\sect\ref{sec:contact_optimisation}). Robot-object contacts for example grasps 5 and 6 are shown in red in \fig\ref{fig:training_b} for estimated models and in \fig\ref{fig:training_c} for optimised models. \fig\ref{fig:contact_optimisation_b} shows that collision model values for estimated model of grasp 5 are very high, which can also be seen in \fig\ref{fig:training_b}. \rev{Contact optimisation not only almost completely removes collisions (collision model value \eq\ref{eq:collision_model_b}), but also increases the contact mass value \eq\ref{eq:contact_optimisation_b} for all training grasp examples (\fig\ref{fig:contact_optimisation_a}).}

\newcommand{\includegraphicstrainingexamples}[2]{\begin{tikzpicture}\node[inner sep=0] (image) at (0,0) {\includegraphics[height=2.0cm]{resources/training-ar/lo/#1.jpg}};\node[font=\bfseries\sffamily, white] at (-0.45,-0.7) {#2};\end{tikzpicture}}
\newcommand{\includegraphicstrainingmodels}[2]{\begin{tikzpicture}\node[inner sep=0] (image) at (0,0) {\includegraphics[height=1.95cm, trim={1.0cm 2.0cm 1.0cm 2.0cm},clip]{resources/training-vr/lo/#1.jpg}};\node[font=\bfseries\sffamily, black] at (-0.5,-0.7) {#2};\end{tikzpicture}}

\begin{figure}[ht]

\begin{center}
\begin{tabular}{cc}%trim=left botm right top
\hspace*{-0.5cm}
\begin{tabular}{c}%trim=left botm right top
\begin{tabular}{cccc}%trim=left botm right top
\begin{tabular}{c}
\hspace*{-0.5cm}\includegraphicstrainingexamples{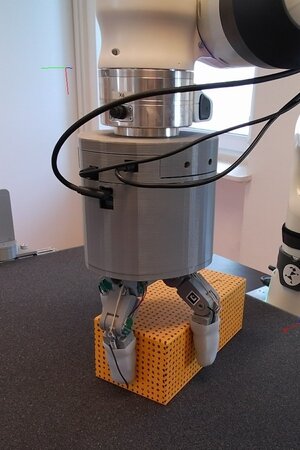}{1}\\
\hspace*{-0.5cm}\includegraphicstrainingexamples{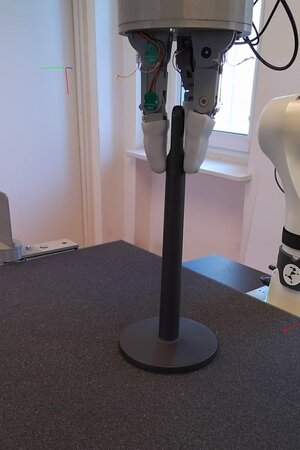}{5}
\end{tabular}&
\begin{tabular}{c}
\hspace*{-0.8cm}\includegraphicstrainingexamples{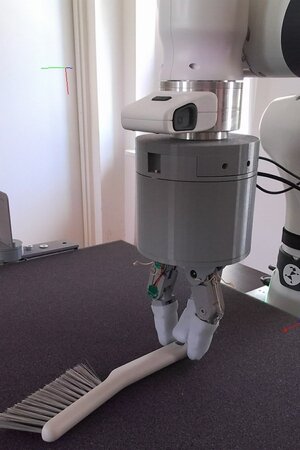}{2}\\
\hspace*{-0.8cm}\includegraphicstrainingexamples{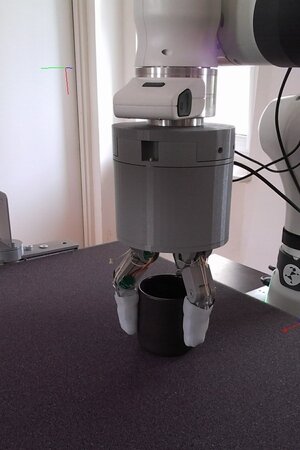}{6}
\end{tabular}&
\begin{tabular}{c}
\hspace*{-0.8cm}\includegraphicstrainingexamples{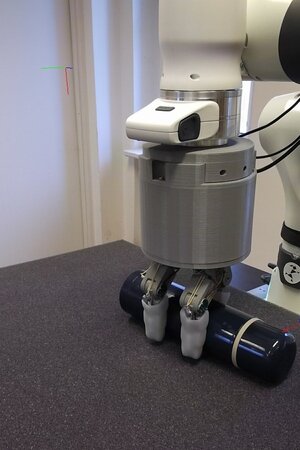}{3}\\
\hspace*{-0.8cm}\includegraphicstrainingexamples{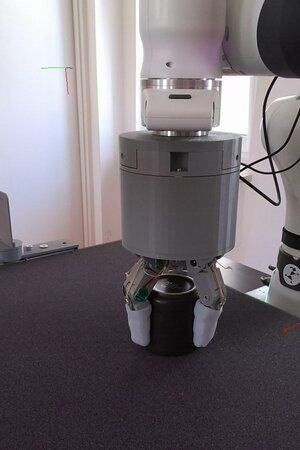}{7}
\end{tabular}&
\begin{tabular}{c}
\hspace*{-0.8cm}\includegraphicstrainingexamples{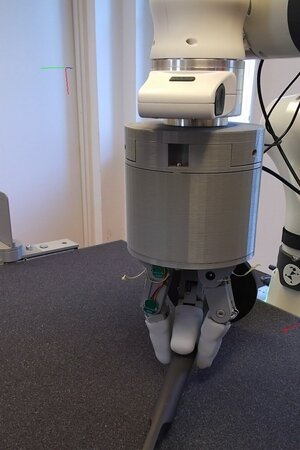}{4}\\
\hspace*{-0.8cm}\includegraphicstrainingexamples{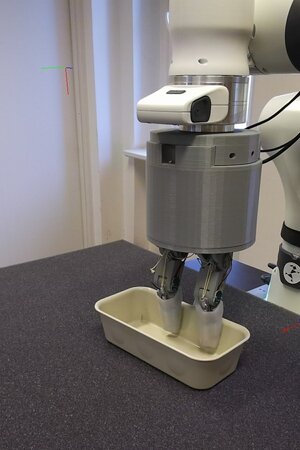}{8}
\end{tabular}
\end{tabular}
\\\hspace*{-1.0cm}\begin{minipage}{0cm}\subcaption{}\label{fig:training_a}\end{minipage}
\end{tabular}

\begin{tabular}{c}\hspace*{-0.3cm}
\begin{tabular}{c}
\hspace*{-0.7cm}\includegraphicstrainingmodels{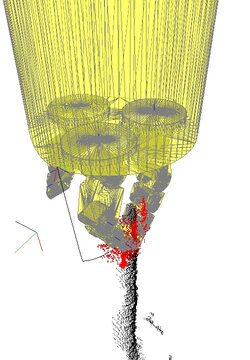}{5}\\
\hspace*{-0.7cm}\includegraphicstrainingmodels{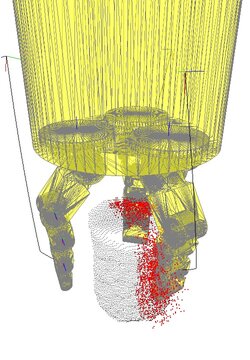}{6}
\end{tabular}
\\\hspace*{-1.2cm}\begin{minipage}{0cm}\subcaption{}\label{fig:training_b}\end{minipage}
\end{tabular}

\begin{tabular}{c}\hspace*{-0.3cm}
\begin{tabular}{c}
\hspace*{-0.5cm}\includegraphicstrainingmodels{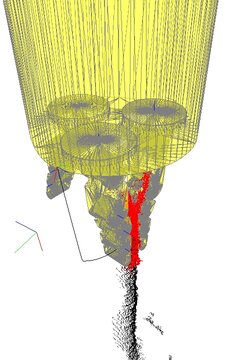}{5}\\
\hspace*{-0.5cm}\includegraphicstrainingmodels{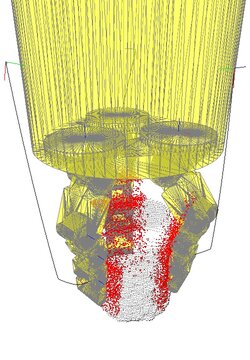}{6}
\end{tabular}
\\\hspace*{-1.2cm}\begin{minipage}{0cm}\subcaption{}\label{fig:training_c}\end{minipage}
\end{tabular}

\end{tabular}
\end{center}\vspace*{-0.4cm}

\caption{Grasp training examples 1-8 (a), estimated (b) and optimised (c) contact models for training example 5 and 6.}
\label{fig:training}
\vspace*{-0.4cm}
\end{figure}

\subsubsection{Inference and execution}
For tests we used 42 IKEA objects at semi-random poses on the table, creating in this way 128 object-pose pairs for grasping. We compared both estimated and optimised models, each time replicating exactly the same object-pose pair, but capturing the test point clouds independently. The robot performed a total of 256 grasp trials fully autonomously, selecting the first most likely grasp that is kinematically feasible, taking into account predefined obstacles in the form of a table, robot, cage and cameras.

\tab\ref{tab:results} shows that optimized models with 80\% success rate perform significantly better than estimated contact models with a success rate of 57\%. 
%The results show that grasp \#2 was never selected and was dominated by the similar grasp \#4, with a more common rounded local shape. 
\rev{Grasps \#2 and \#4 have similar gripper configurations, but different surface properties at contacts (rectangular vs. rounded). This affected corresponding query densities, so that grasp \#2 was never selected and was dominated by the similar grasp \#4, with a more common rounded local shape.}
Moreover, estimated models often chose pinch grasps \#4 and \#5 with very poor success rates. The optimized models significantly improved their reliability, but also reduced their selection probability.

\fig\ref{fig:results} illustrates a few grasping results, including failures. The most common cause of failure for both models was an inappropriate (too weak) grasp type for a given object. Before executing each grasp, the reconfiguration planner \sect\ref{sec:reconfiguration_planning} was run to generate trajectories \eq\ref{eq:reconfig_interpol} to reconfigure the \rp joints for the grasp selected by the grasp inference algorithm \sect\ref{sec:contact_models}, if required. An example reconfiguration trajectory execution is shown in \fig\ref{fig:reconfig}.

\rev{We further validated our algorithm with optimized models on the YCB~\cite{calli2015ycb} dataset, with a success rate of 87\% on 120 grasps following the protocol from \cite{bekiroglu2019benchmarking}. \cite{bekiroglu2019benchmarking} shows 95\% rate on this benchmark, however it relies on 4 views and a more powerful industry-grade parallel-jaw gripper (see also \tab\ref{tab:comparison}).}

\newcommand{\includegraphicsresultsplan}[1]{\includegraphics[height=2.1cm, trim={3.5cm 4.5cm 9cm 0cm},clip]{resources/test/lo/#1.jpg}}
\newcommand{\includegraphicsresultsplanindex}[2]{\begin{tikzpicture}\node[inner sep=0] (image) at (0,0) {\includegraphics[height=1.95cm, trim={3.5cm 4cm 10.0cm 0cm},clip]{resources/test/lo/#1.jpg}};\node[font=\bfseries\sffamily, white] at (+0.4,-0.7) {#2};\end{tikzpicture}}
\newcommand{\includegraphicsresultsgrasp}[1]{\includegraphics[height=2.1cm, trim={4cm 1cm 4.5cm 0},clip]{resources/test/lo/#1.jpg}}

\begin{table}[ht]
\rev{
\begin{center}
\footnotesize
\hspace*{-0.03cm}
\begin{tabular}{|p{0.06\textwidth}|p{0.01\textwidth} p{0.01\textwidth} p{0.02\textwidth} p{0.02\textwidth} p{0.02\textwidth} p{0.021\textwidth} p{0.02\textwidth} p{0.02\textwidth} p{0.043\textwidth}|}
\hline
\hspace*{-0.1cm}Models & 1 & 2 & 3 & 4 & 5 & 6 & 7 & 8 & \textbf{All} \\
\hline
\hspace*{-0.1cm}Estimated & 5/7 & - & 29/37 & 7/25 & 2/19 & 14/16 & 6/8 & 10/16 & \textbf{73/128} \\
\hspace*{-0.1cm}Optimised & 2/2 & - & 45/58 & 2/2 & 4/7 & 13/13 & 12/12 & 24/34 & \textbf{102/128} \\
\hspace*{-0.1cm}YCB+Opt. & - & - & 68/81 & - & - & 21/22 & 11/11 & 4/6 & \textbf{104/120} \\
\hline
\end{tabular}
\end{center}
\vspace*{-0.3cm}\caption{\rev{Number of successful/total grasps by training example 1-8 and in overall, for estimated and optimised models.}}
\label{tab:results}
\vspace*{-0.3cm}
}
\end{table}

\begin{figure}[ht]\vspace*{-0.2cm}
\begin{center}
\includegraphicsresultsplanindex{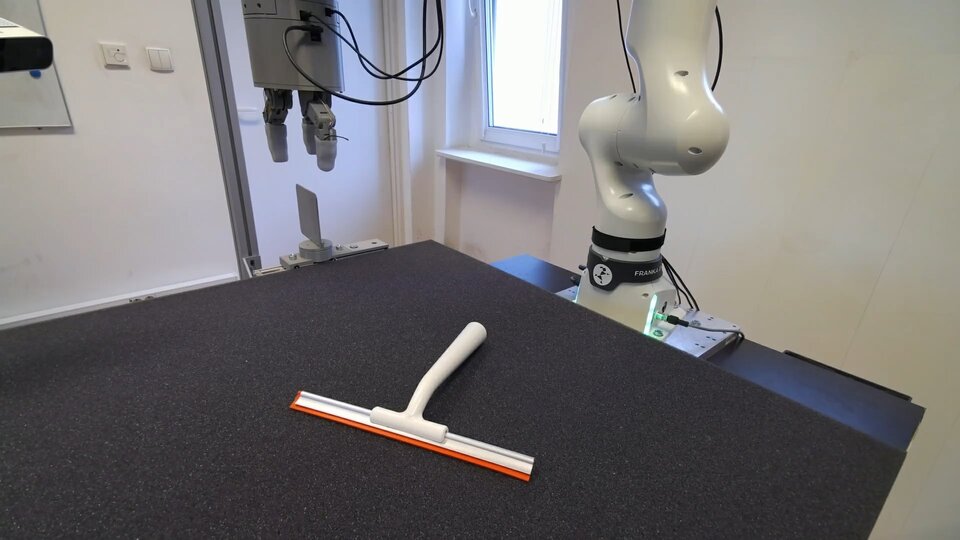}{1}
\hspace*{-0.08cm}\includegraphicsresultsplanindex{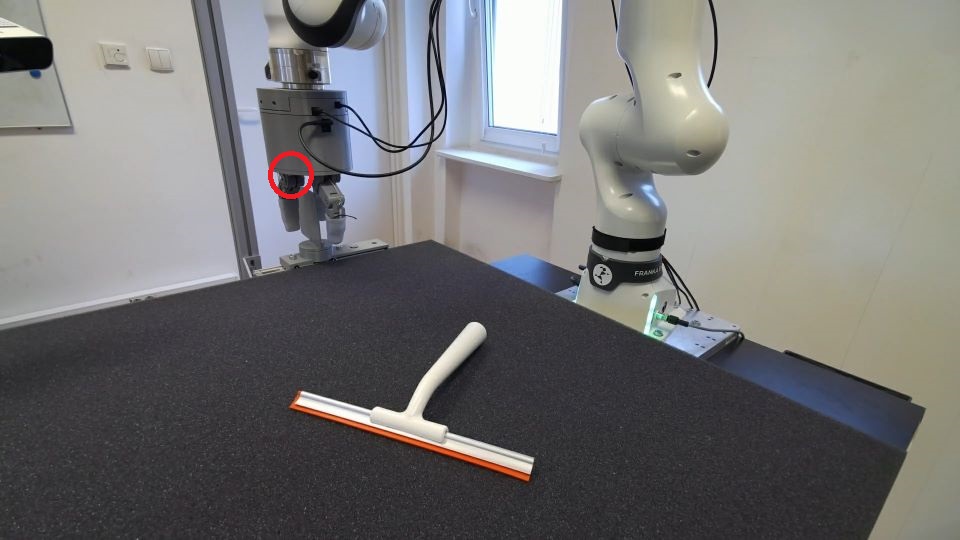}{2}
\hspace*{-0.08cm}\includegraphicsresultsplanindex{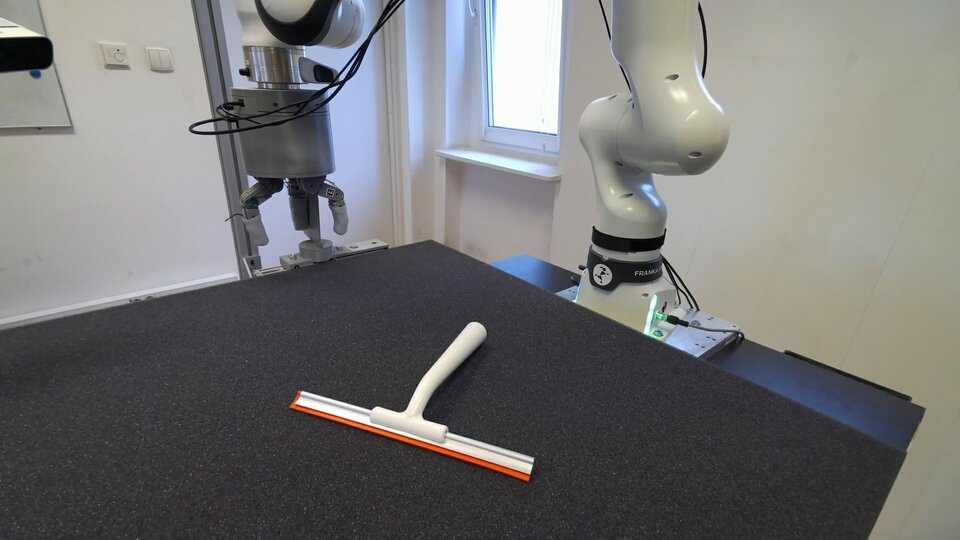}{3}
\hspace*{-0.08cm}\includegraphicsresultsplanindex{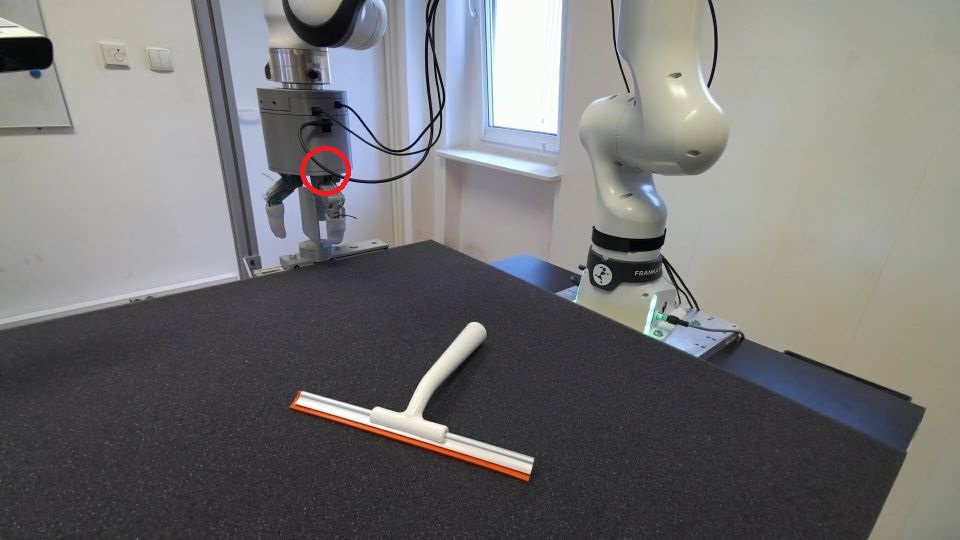}{4}
\hspace*{-0.08cm}\includegraphicsresultsplanindex{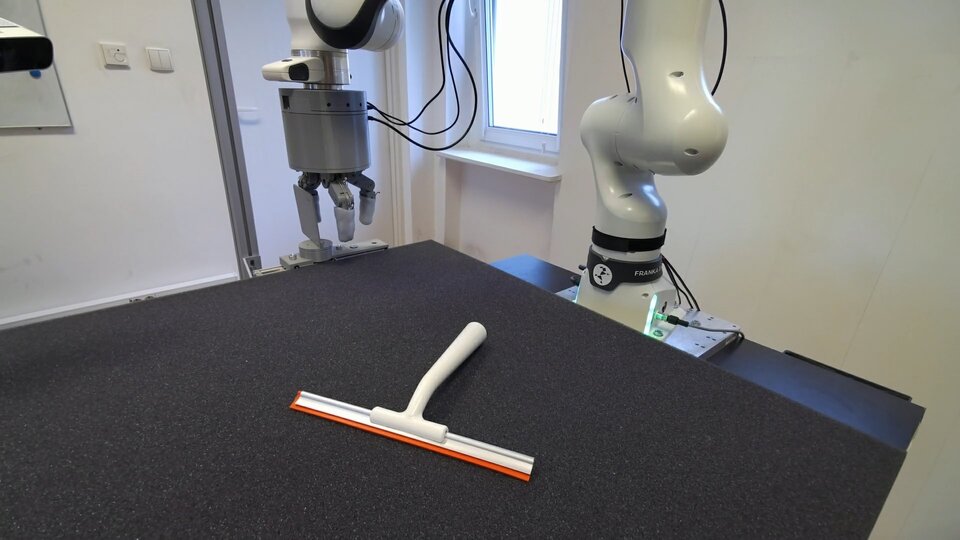}{5}
\hspace*{-0.08cm}\includegraphicsresultsplanindex{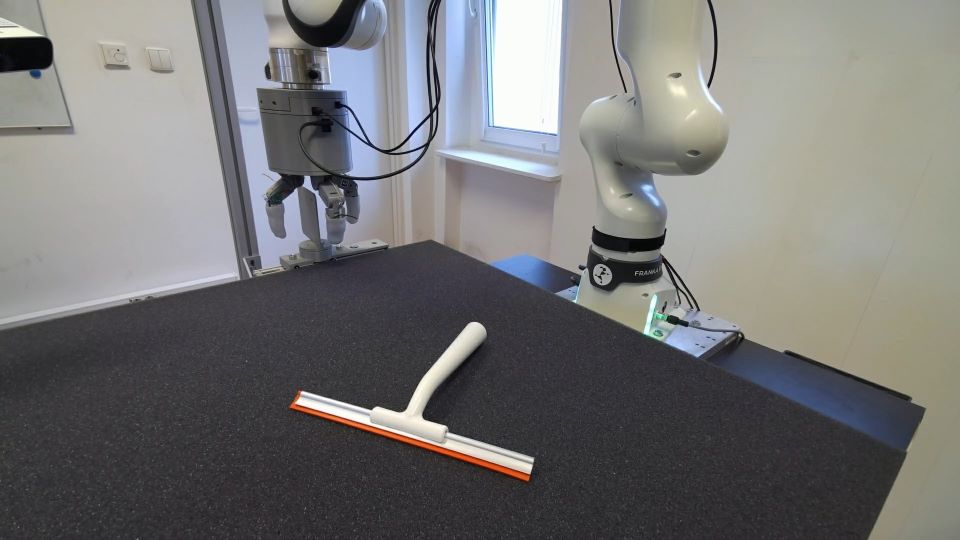}{6}
\hspace*{-0.08cm}\includegraphicsresultsplanindex{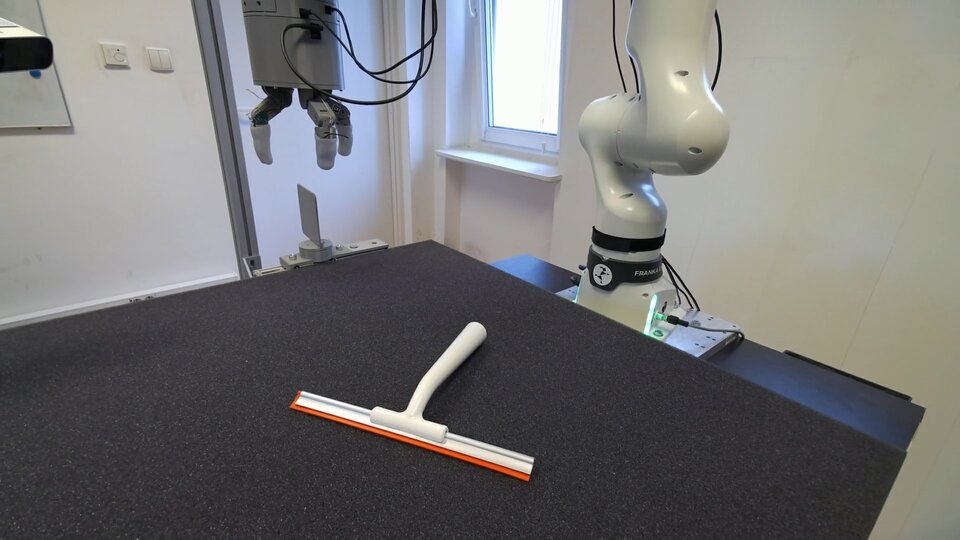}{7}
\end{center}
\vspace*{-0.3cm}\caption{\rev{To reconfigure 2 \rp joints from 0\textdegree~to 90\textdegree, our planner creates a sequence of 6 trajectories, where the end waypoint of each trajectory becomes the starting waypoint of the next trajectory in the sequence (except the last one). 1-2 approach trajectory, 2-3 reconfiguration of the 1st \rp joint (red circle), 3-4 rollback of 2-3 trajectory, 4-5 reconfiguration of the 2nd \rp joint (red circle, finger at the back), 5-6 rollback of 4-5 trajectory and 6-7 rollback of 1-2 trajectory.}}
\label{fig:reconfig}
\vspace*{-0.4cm}
\end{figure}

\begin{figure}[ht]

\begin{center}
\vspace*{0.1cm}
\begin{tabular}{cc}%trim=left botm right top
\hspace*{-0.3cm}
\begin{tabular}{c}%trim=left botm right top
\begin{tabular}{cc}%trim=left botm right top
\begin{tabular}{c}
\hspace*{-0.6cm}\includegraphicsresultsgrasp{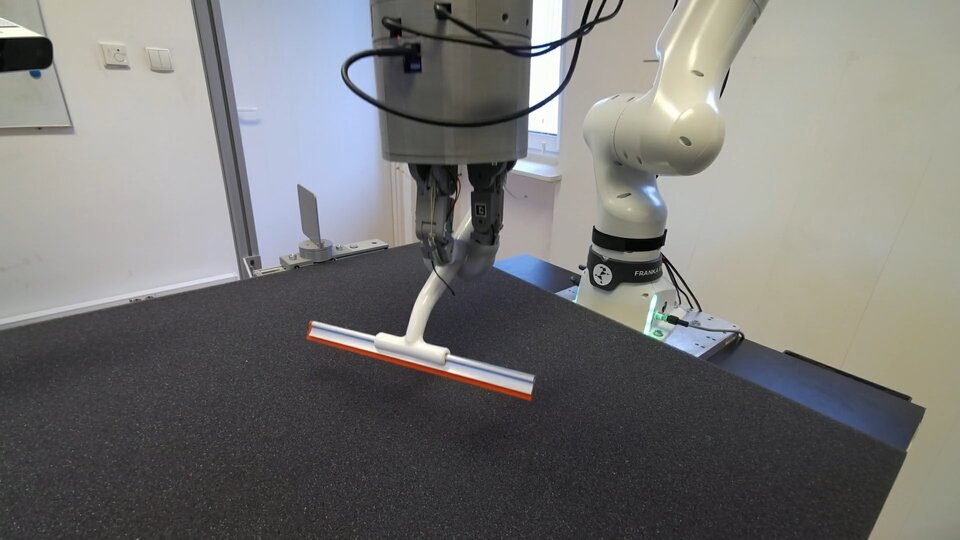}\\
\hspace*{-0.6cm}\includegraphicsresultsgrasp{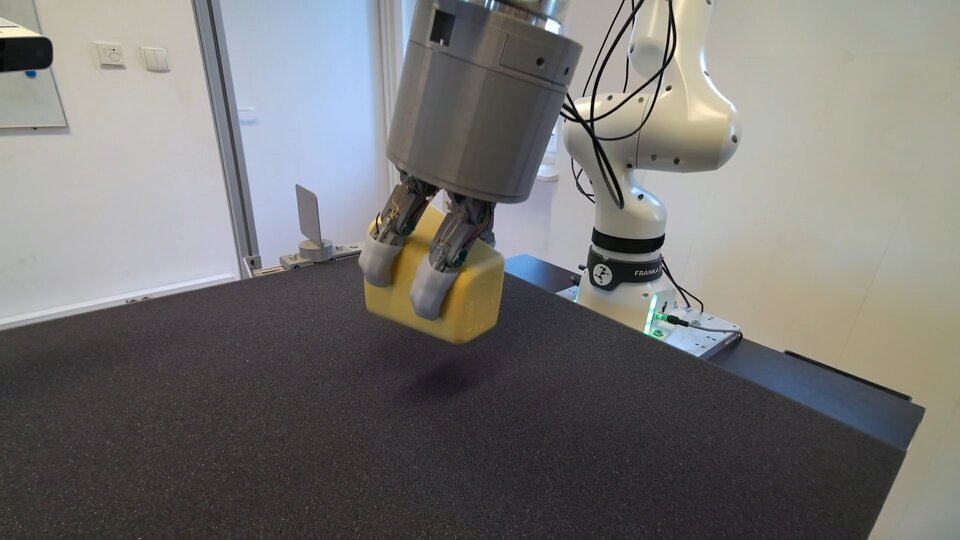}\\
\hspace*{-0.6cm}\includegraphicsresultsgrasp{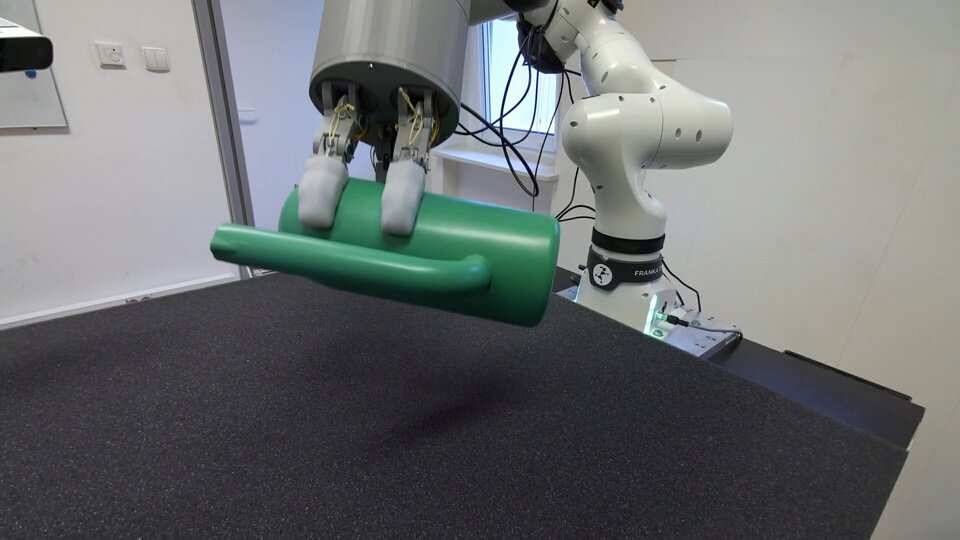}
\end{tabular}&
\begin{tabular}{c}
\hspace*{-0.8cm}\includegraphicsresultsgrasp{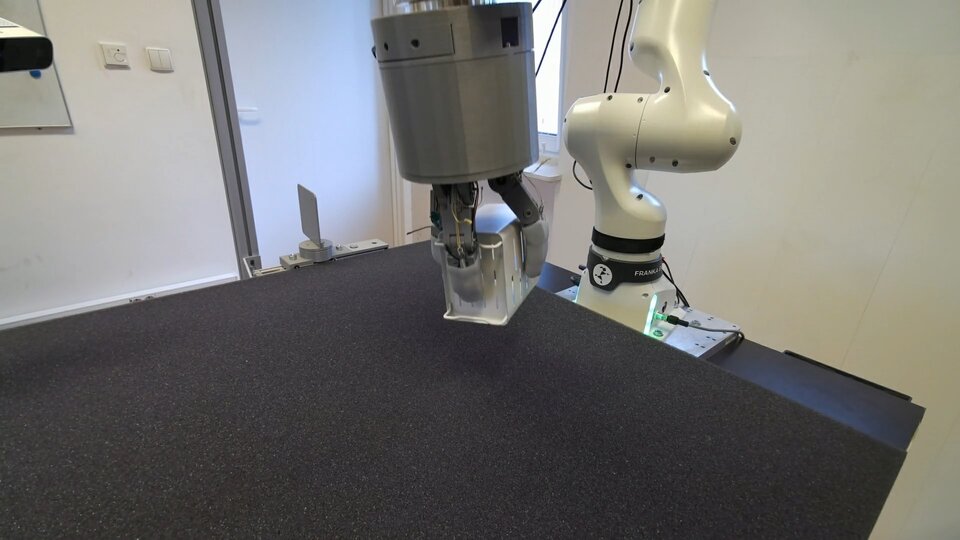}\\
\hspace*{-0.8cm}\includegraphicsresultsgrasp{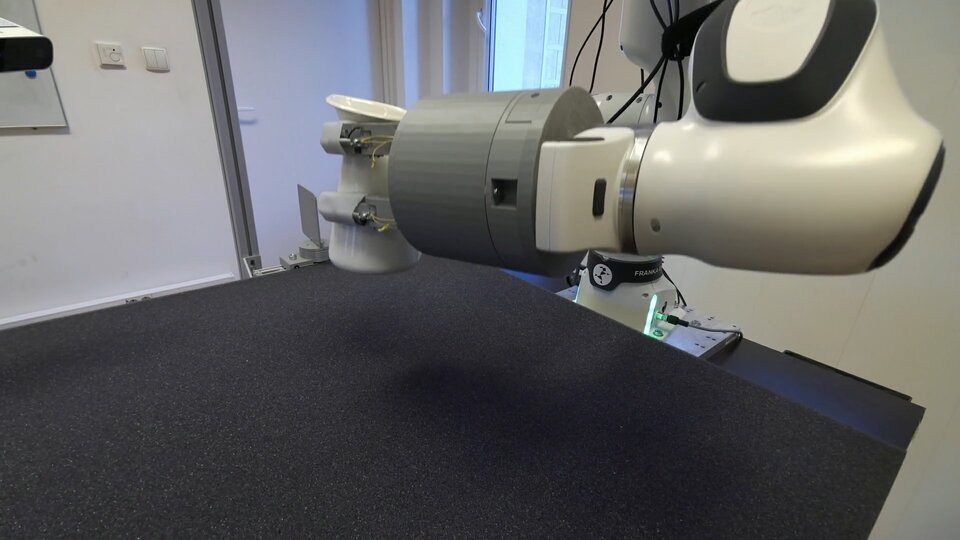}\\
\hspace*{-0.8cm}\includegraphicsresultsgrasp{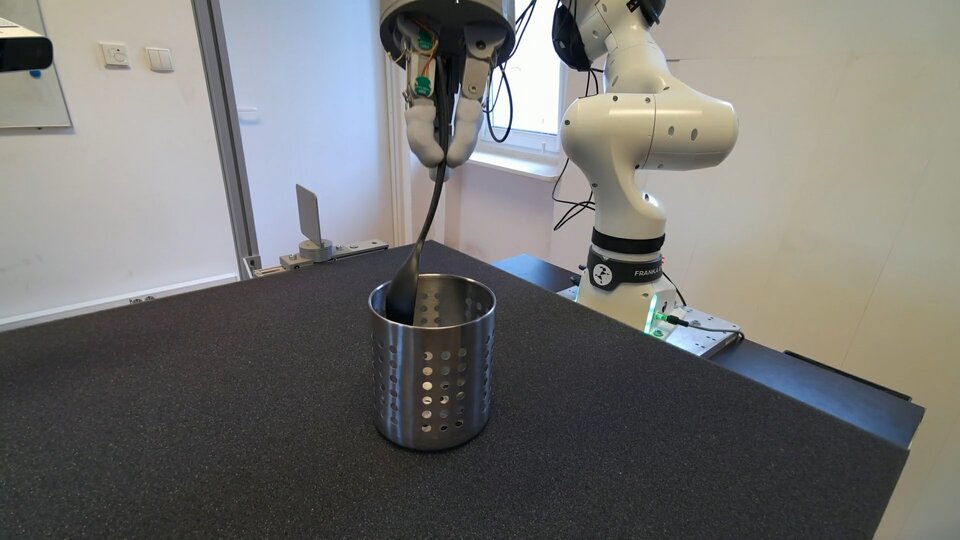}
\end{tabular}
\end{tabular}
%\\\hspace*{-0.9cm}\begin{minipage}{0cm}\subcaption{}\label{fig:results_a}\end{minipage}
\end{tabular}

\begin{tabular}{c}%trim=left botm right top
\begin{tabular}{cc}%trim=left botm right top
\hspace*{-0.6cm}
\begin{tabular}{c}
\hspace*{-0.6cm}\includegraphicsresultsgrasp{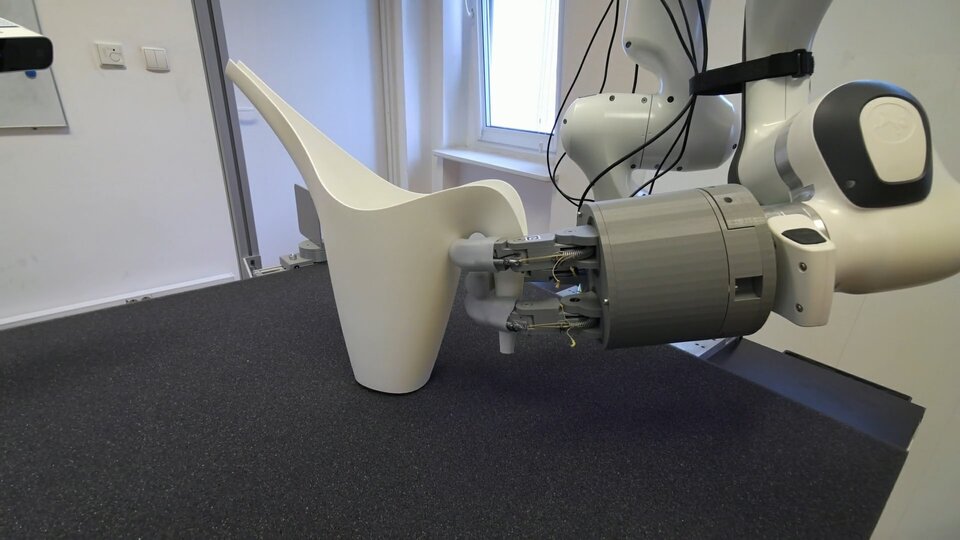}\\
\hspace*{-0.6cm}\includegraphicsresultsgrasp{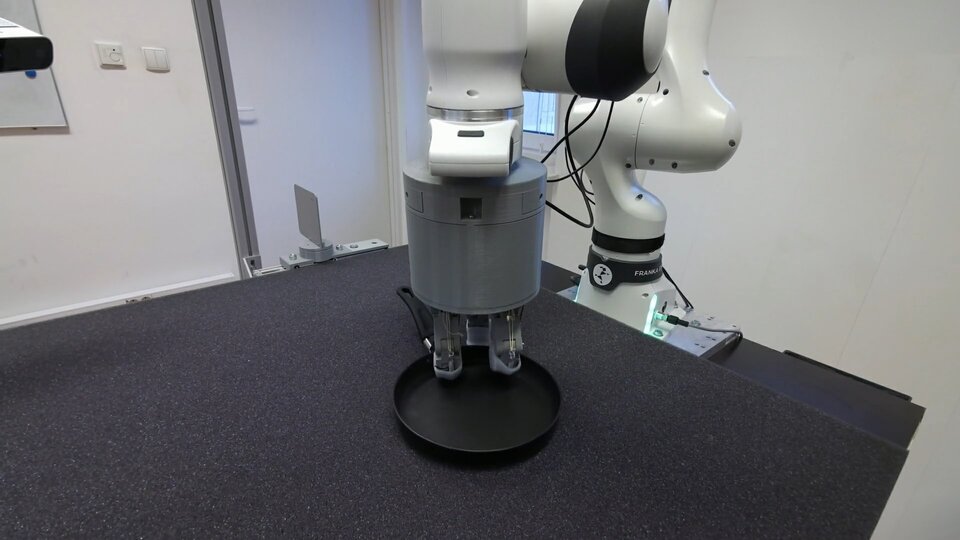}\\
\hspace*{-0.6cm}\includegraphicsresultsgrasp{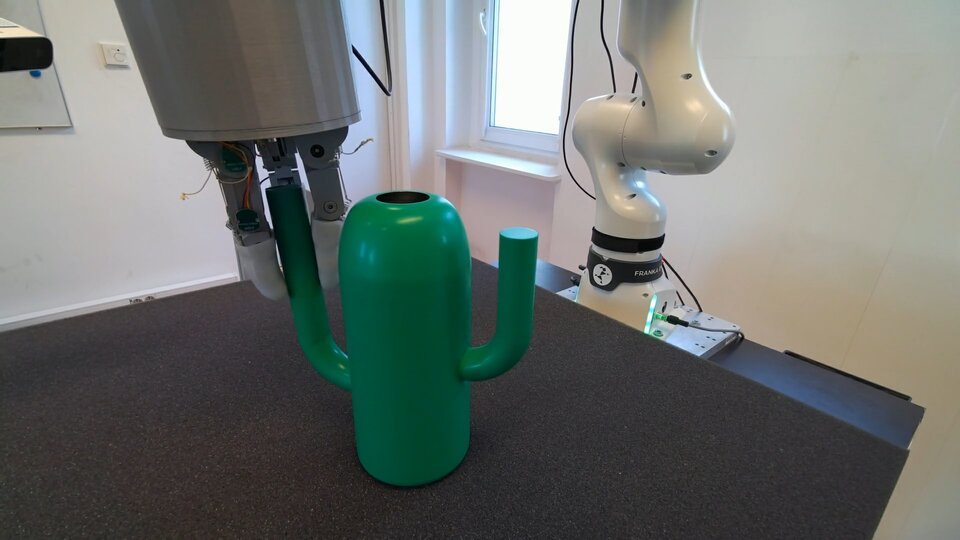}
\end{tabular}&
\begin{tabular}{c}
\hspace*{-0.8cm}\includegraphicsresultsgrasp{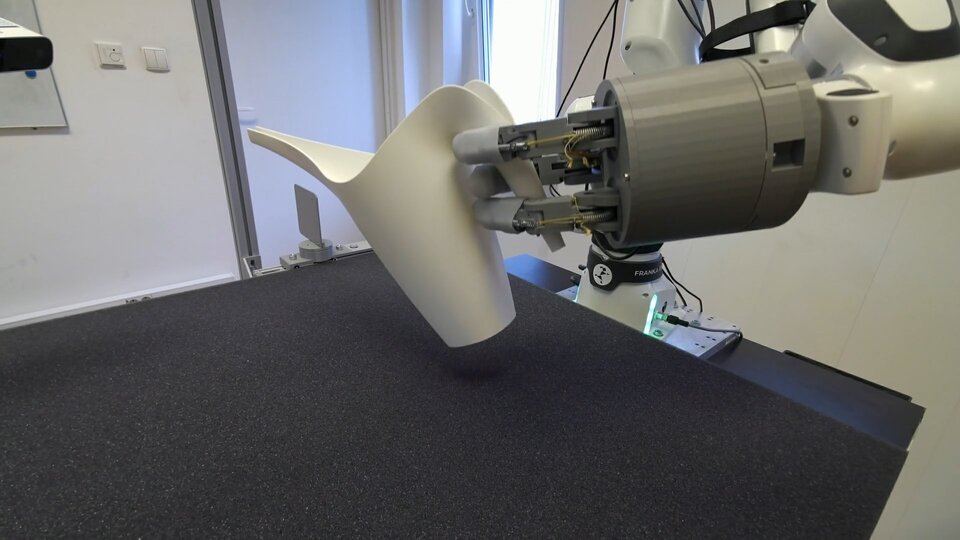}\\
\hspace*{-0.8cm}\includegraphicsresultsgrasp{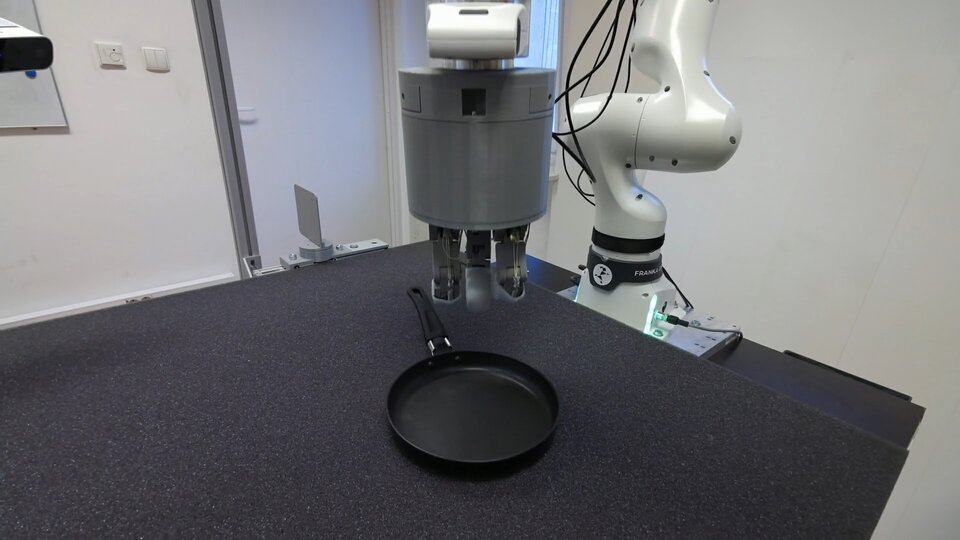}\\
\hspace*{-0.8cm}\includegraphicsresultsgrasp{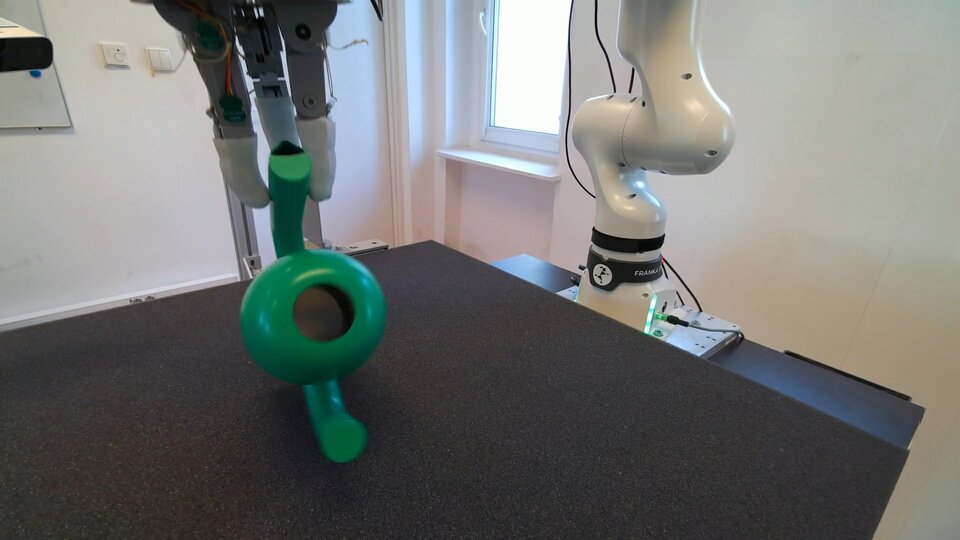}
\end{tabular}
\end{tabular}
%\\\hspace*{-1.3cm}\begin{minipage}{0cm}\subcaption{}\label{fig:results_b}\end{minipage}
\end{tabular}
\end{tabular}
\end{center}\vspace*{-0.4cm}

\caption{Successful grasps (left panel), and 3 example trajectories - one successful and two failures (right panel).}
\label{fig:results}\vspace*{-0.5cm}
\end{figure}

\section{Conclusions}

% Remarks by PS
%\subsubsection{Robotic gripper}
%The addition of \rp joints in the gripper provides two extra DoF, enabling the fingers to spread sideways during pre-grasp configuration and lock during grasping. This enhances dexterous grasping by allowing the fingers to conform to various object shapes and sizes, making the gripper more versatile. The design facilitates both parallel and enveloping grasps, enabling it to handle lightweight objects from the sides without pushing them away, and to securely grasp small objects with parallel fingers. Moreover, the \rp joint compliance in the second joint allows the last three links to move bidirectionally under external forces, absorbing impacts and reducing the risk of collision damage during manipulation tasks. This not only increases the gripper's reliability but also ensures it is capable of handling complex environments and various grasping scenarios efficiently.

We proposed an underactuated three-finger gripper with two reconfigurable passive joints (\rp joint), enabling the fingers to flexibly re-configure during pre-grasp and lock during grasping. We also simplified the enveloping/parallel mechanism which reduces the design complexity. To execute grasps with the developed gripper, we designed a planning algorithm for underactuated grippers, which learns dexterous grasps from single examples exploiting also \rp joints reconfigurability. Kinaesthetic contact optimization has also been included to increase grasp performance. Experimental validation  on 250 grasps performed on 42 IKEA objects and \rev{the YCB object dataset} showed that the combination of the proposed gripper and planner can effectively perform grasps, handling complex environments and various grasping scenarios.

%\subsubsection*{Future work} will involve applications of RP-joints in prosthetic hands and further work on manipulation and grasping algorithms with underactuated hands to improve its performance in industrial settings.
\rev{\textit{Future work} will include applications of \rp joints in prosthetic hands, as well as further work on manipulation and grasping algorithms for underactuated hands to improve their performance in industrial settings.}

%\appendices

%\section*{Acknowledgment}
\vspace*{-0.2cm}
\bibliographystyle{IEEEtran}
\bibliography{IEEEabrv,bibliography}

\begin{thebibliography}{10}
\providecommand{\url}[1]{#1}
\csname url@samestyle\endcsname
\providecommand{\newblock}{\relax}
\providecommand{\bibinfo}[2]{#2}
\providecommand{\BIBentrySTDinterwordspacing}{\spaceskip=0pt\relax}
\providecommand{\BIBentryALTinterwordstretchfactor}{4}
\providecommand{\BIBentryALTinterwordspacing}{\spaceskip=\fontdimen2\font plus
\BIBentryALTinterwordstretchfactor\fontdimen3\font minus
  \fontdimen4\font\relax}
\providecommand{\BIBforeignlanguage}[2]{{%
\expandafter\ifx\csname l@#1\endcsname\relax
\typeout{** WARNING: IEEEtran.bst: No hyphenation pattern has been}%
\typeout{** loaded for the language `#1'. Using the pattern for}%
\typeout{** the default language instead.}%
\else
\language=\csname l@#1\endcsname
\fi
#2}}
\providecommand{\BIBdecl}{\relax}
\BIBdecl

\bibitem{bonilla2014grasping}
M.~Bonilla \emph{et~al.}, ``Grasping with soft hands,'' in \emph{2014 IEEE-RAS
  Int. Conf. on Humanoid Robots}.\hskip 1em plus 0.5em minus 0.4em\relax IEEE,
  2014, pp. 581--587.

\bibitem{tavakoli2013flexirigid}
M.~Tavakoli \emph{et~al.}, ``Flexirigid, a novel two phase flexible gripper,''
  in \emph{2013 IEEE/RSJ Int. Conf. on Intelligent Robots and Systems}.\hskip
  1em plus 0.5em minus 0.4em\relax IEEE, 2013, pp. 5046--5051.

\bibitem{Hsu2017}
J.~Hsu \emph{et~al.}, ``Self-locking underactuated mechanism for robotic
  gripper,'' in \emph{IEEE Int. Conf. on Advanced Intelligent Mechatronics
  (AIM)}, 2017, pp. 620--627.

\bibitem{calli2015ycb}
B.~Calli \emph{et~al.}, ``The ycb object and model set: Towards common
  benchmarks for manipulation research,'' in \emph{2015 Int. Conf. on advanced
  robotics (ICAR)}.\hskip 1em plus 0.5em minus 0.4em\relax IEEE, 2015, pp.
  510--517.

\bibitem{shintake2018soft}
J.~Shintake \emph{et~al.}, ``Soft robotic grippers,'' \emph{Advanced
  materials}, vol.~30, no.~29, p. 1707035, 2018.

\bibitem{deimel2013compliant}
R.~Deimel \emph{et~al.}, ``A compliant hand based on a novel pneumatic
  actuator,'' in \emph{2013 IEEE Int. Conf. on Robotics and Automation}.\hskip
  1em plus 0.5em minus 0.4em\relax IEEE, 2013, pp. 2047--2053.

\bibitem{deimel2016novel}
------, ``A novel type of compliant and underactuated robotic hand for
  dexterous grasping,'' \emph{The Int. J. of Robotics Research}, vol.~35, no.
  1-3, pp. 161--185, 2016.

\bibitem{glick2018soft}
P.~Glick \emph{et~al.}, ``A soft robotic gripper with gecko-inspired
  adhesive,'' \emph{IEEE Robotics and Automation Letters}, vol.~3, no.~2, pp.
  903--910, 2018.

\bibitem{odhner2014compliant}
L.~U. Odhner \emph{et~al.}, ``A compliant, underactuated hand for robust
  manipulation,'' \emph{The Int. J. of Robotics Research}, vol.~33, no.~5, pp.
  736--752, 2014.

\bibitem{liu2023hybrid}
F.~Liu \emph{et~al.}, ``Hybrid robotic grasping with a soft multimodal gripper
  and a deep multistage learning scheme,'' \emph{IEEE Trans. on Robotics},
  2023.

\bibitem{della2018toward}
C.~Della~Santina \emph{et~al.}, ``Toward dexterous manipulation with augmented
  adaptive synergies: The pisa/iit softhand 2,'' \emph{IEEE Trans. on
  Robotics}, vol.~34, no.~5, pp. 1141--1156, 2018.

\bibitem{angelini2020softhandler}
F.~Angelini \emph{et~al.}, ``Softhandler: An integrated soft robotic system for
  handling heterogeneous objects,'' \emph{IEEE Robotics \& Automation
  Magazine}, vol.~27, no.~3, pp. 55--72, 2020.

\bibitem{Lu2021}
Q.~Lu \emph{et~al.}, ``Systematic object-invariant in-hand manipulation via
  reconfigurable underactuation: Introducing the {RUTH} gripper,'' \emph{The
  Int. J. of Robotics Research}, vol.~40, no. 12--14, pp. 1402--1418, 2021.

\bibitem{Chappel2023}
D.~Chappell \emph{et~al.}, ``The hydra hand: A mode-switching underactuated
  gripper with precision and power grasping modes,'' \emph{IEEE Robotics and
  Automation Letters}, vol.~8, no.~11, pp. 7599--7606, 2023.

\bibitem{Plooij2015}
M.~Plooij \emph{et~al.}, ``Lock your robot: A review of locking devices in
  robotics,'' \emph{IEEE Robotics \& Automation Magazine}, vol.~22, no.~1, pp.
  106--117, 2015.

\bibitem{Hermann2019}
K.~Hermann \emph{et~al.}, ``A joint-selective robotic gripper with actuation
  mode switching,'' in \emph{IEEE 15th Int. Conf. on Automation Science and
  Engineering (CASE)}, 2019, pp. 1532--1539.

\bibitem{mitsui2013under}
K.~Mitsui \emph{et~al.}, ``An under-actuated robotic hand for multiple
  grasps,'' in \emph{2013 IEEE/RSJ Int. Conf. on Intelligent Robots and
  Systems}.\hskip 1em plus 0.5em minus 0.4em\relax IEEE, 2013, pp. 5475--5480.

\bibitem{peerdeman2013development}
B.~Peerdeman \emph{et~al.}, ``Development of underactuated prosthetic fingers
  with joint locking and electromyographic control,'' \emph{Mechanical
  engineering research}, vol.~3, no.~1, p. 130, 2013.

\bibitem{newbury2023deep}
R.~Newbury \emph{et~al.}, ``Deep learning approaches to grasp synthesis: A
  review,'' \emph{IEEE Trans. on Robotics}, 2023.

\bibitem{billard2019trends}
A.~Billard \emph{et~al.}, ``Trends and challenges in robot manipulation,''
  \emph{Science}, vol. 364, no. 6446, p. eaat8414, 2019.

\bibitem{bicchi2000robotic}
A.~Bicchi \emph{et~al.}, ``Robotic grasping and contact: A review,'' in
  \emph{Proceedings 2000 ICRA. Millennium conference. IEEE Int. Conf. on
  robotics and automation. Symposia proceedings (Cat. No. 00CH37065)},
  vol.~1.\hskip 1em plus 0.5em minus 0.4em\relax IEEE, 2000, pp. 348--353.

\bibitem{pollayil2022sequential}
G.~J. Pollayil \emph{et~al.}, ``Sequential contact-based adaptive grasping for
  robotic hands,'' \emph{The Int. J. of Robotics Research}, vol.~41, no.~5, pp.
  543--570, 2022.

\bibitem{kleeberger2020survey}
K.~Kleeberger \emph{et~al.}, ``A survey on learning-based robotic grasping,''
  \emph{Current Robotics Reports}, vol.~1, pp. 239--249, 2020.

\bibitem{mahler2019learning}
J.~Mahler \emph{et~al.}, ``Learning ambidextrous robot grasping policies,''
  \emph{Science Robotics}, vol.~4, no.~26, p. eaau4984, 2019.

\bibitem{levine2018learning}
S.~Levine \emph{et~al.}, ``Learning hand-eye coordination for robotic grasping
  with deep learning and large-scale data collection,'' \emph{The Int. J. of
  robotics research}, vol.~37, no. 4-5, pp. 421--436, 2018.

\bibitem{ten2017grasp}
A.~Ten~Pas \emph{et~al.}, ``Grasp pose detection in point clouds,'' \emph{The
  Int. J. of Robotics Research}, vol.~36, no. 13-14, pp. 1455--1473, 2017.

\bibitem{ainetter2021end}
S.~Ainetter \emph{et~al.}, ``End-to-end trainable deep neural network for
  robotic grasp detection and semantic segmentation from rgb,'' in \emph{2021
  IEEE Int. Conf. on Robotics and Automation (ICRA)}.\hskip 1em plus 0.5em
  minus 0.4em\relax IEEE, 2021, pp. 13\,452--13\,458.

\bibitem{Aktas2022Deep}
U.~R. Akta\c{s} \emph{et~al.}, ``Deep dexterous grasping of novel objects from
  a single view,'' \emph{Int. J. of Humanoid Robotics}, vol.~19, no.~02, 2022.

\bibitem{urain2023se}
J.~Urain \emph{et~al.}, ``Se (3)-diffusionfields: Learning smooth cost
  functions for joint grasp and motion optimization through diffusion,'' in
  \emph{2023 IEEE Int. Conf. on Robotics and Automation (ICRA)}.\hskip 1em plus
  0.5em minus 0.4em\relax IEEE, 2023, pp. 5923--5930.

\bibitem{kopicki2014learning}
M.~Kopicki \emph{et~al.}, ``Learning dexterous grasps that generalise to novel
  objects by combining hand and contact models,'' in \emph{{IEEE} Int. Conf. on
  Robotics and Automation (ICRA)}.\hskip 1em plus 0.5em minus 0.4em\relax
  {IEEE}, 2014, pp. 5358--5365.

\bibitem{kopicki2016oneshot}
------, ``One-shot learning and generation of dexterous grasps for novel
  objects,'' \emph{The Int. J. of Robotics Research}, vol.~35, no.~8, pp.
  959--976, 2016.

\bibitem{kopicki2019learning}
------, ``Learning better generative models for dexterous, single-view grasping
  of novel objects,'' \emph{The Int. J. of Robotics Research}, vol.~38, no.
  10-11, pp. 1246--1267, 2019.

\bibitem{della2019learning}
C.~Della~Santina \emph{et~al.}, ``Learning from humans how to grasp: a
  data-driven architecture for autonomous grasping with anthropomorphic soft
  hands,'' \emph{IEEE Robotics and Automation Letters}, vol.~4, no.~2, pp.
  1533--1540, 2019.

\bibitem{palleschi2023grasp}
A.~Palleschi \emph{et~al.}, ``Grasp it like a pro 2.0: A data-driven approach
  exploiting basic shape decomposition and human data for grasping unknown
  objects,'' \emph{IEEE Trans. on Robotics}, 2023.

\bibitem{bekiroglu2019benchmarking}
Y.~Bekiroglu \emph{et~al.}, ``Benchmarking protocol for grasp planning
  algorithms,'' \emph{IEEE Robotics and Automation Letters}, vol.~5, no.~2, pp.
  315--322, 2019.

\bibitem{ciocarlie2014velo}
M.~Ciocarlie \emph{et~al.}, ``The velo gripper: A versatile single-actuator
  design for enveloping, parallel and fingertip grasps,'' \emph{The Int. J. of
  Robotics Research}, vol.~33, no.~5, pp. 753--767, 2014.

\bibitem{golem_short}
M.~Kopicki \emph{et~al.}, ``Golem - robot control, planning and learning
  framework,'' 2009.

\bibitem{fisher1953dispersion}
R.~Fisher, ``Dispersion on a sphere,'' \emph{Proceedings of the Royal Society
  of London. Series A. Mathematical and Physical Sciences}, vol. 217, no. 1130,
  p. 295, 1953.

\bibitem{kopicki2016learning}
M.~Kopicki \emph{et~al.}, ``Learning and inference of dexterous grasps for
  novel objects with underactuated hands,'' \emph{CoRR}, vol. abs/1609.07592,
  2016.

\bibitem{Ko2020}
T.~Ko, ``A tendon-driven robot gripper with passively switchable underactuated
  surface and its physics simulation based parameter optimization,'' \emph{IEEE
  Robotics and Automation Letters}, vol.~5, no.~4, pp. 5002--5009, 2020.

\bibitem{hirose1978development}
S.~Hirose \emph{et~al.}, ``The development of soft gripper for the versatile
  robot hand,'' \emph{Mechanism and machine theory}, vol.~13, no.~3, pp.
  351--359, 1978.

\end{thebibliography}

%\printbibliography

\end{document}